\documentclass[letterpaper]{article}
\usepackage{aaai23} 
\usepackage{times}
\usepackage{helvet} 
\usepackage{courier} 
\usepackage[hyphens]{url} 
\usepackage{graphicx} 
\usepackage{natbib} 
\usepackage{caption} 
\usepackage{algpseudocode}
\usepackage{amsmath}
\usepackage{hyperref}
\usepackage{xcolor,colortbl}
\usepackage{hhline}
\usepackage{amsmath}
\usepackage{multirow}
\usepackage{amsfonts}
\usepackage{caption}
\usepackage{tikz}
\usepackage{pgfplots}
\usepackage{booktabs}
\usepackage{pifont}
\usepackage{float}
\usepackage{tcolorbox}
\usepackage{tabularx}
\usepackage[flushleft]{threeparttable}
\usepackage[T1]{fontenc}
\usepackage[titlenumbered,ruled,noline,linesnumbered]{algorithm2e}
\usepackage{gensymb}
\usepackage{flushend}
\usepackage{makecell}
\usepackage{subfigure}
\usepackage[bottom]{footmisc} 
\usepackage{microtype}
\usepackage{amssymb}

\usepackage{newfloat}
\usepackage{listings}
\DeclareCaptionStyle{ruled}{labelfont=normalfont,labelsep=colon,strut=off} % DO NOT CHANGE THIS
\floatstyle{ruled}
\newfloat{listing}{tb}{lst}{}
\floatname{listing}{Listing}

\usetikzlibrary{calc}
\usetikzlibrary{patterns}
\usetikzlibrary{shapes.multipart, arrows, positioning,shapes.geometric}
\usetikzlibrary{decorations.pathreplacing}
\usetikzlibrary{calc,shapes,decorations,decorations.shapes,positioning,shadows,arrows,decorations.markings,backgrounds,fit}

\makeatletter
\newcommand*{\rom}[1]{\expandafter\@slowromancap\romannumeral #1@}
\makeatother
\renewcommand{\vec}[1]{\mathbf{#1}}
\newcommand{\codename}{{\sc Purifier}\xspace}

\DeclareMathOperator*{\argmax}{arg\,max}

\newcolumntype{Y}{>{\centering\arraybackslash}X}

\makeatletter
\newcommand{\removelatexerror}{\let\@latex@error\@gobble}
\makeatother

\let\emptyset\varnothing

\usepackage{bigfoot}
\DeclareNewFootnote{ANote}[fnsymbol]

\begin{document}
\title{Purifier: Defending Data Inference Attacks via Transforming Confidence Scores}
\author {
    Ziqi Yang\textsuperscript{\rm 1,\rm 2,\rm 3},
    Lijin Wang\textsuperscript{\rm 1},
    Da Yang\textsuperscript{\rm 1},
    Jie Wan\textsuperscript{\rm 1},\\
    Ziming Zhao\textsuperscript{\rm 1},
    Ee-Chien Chang\textsuperscript{\rm 6},
    Fan Zhang\textsuperscript{\rm 1,\rm 4,\rm5}\thanks{Corresponding author},
    Kui Ren \textsuperscript{\rm 1,\rm 2,\rm 3,\rm4}
}
\affiliations {
    \textsuperscript{\rm 1} Zhejiang University\\
    \textsuperscript{\rm 2} ZJU-Hangzhou Global Scientific and Technological Innovation Center, Zhejiang University\\
    \textsuperscript{\rm 3} Key Laboratory of Blockchain and Cyberspace Governance of Zhejiang Province\\
    \textsuperscript{\rm 4} Jiaxing Research Institute, Zhejiang University\\
    \textsuperscript{\rm 5} Zhengzhou Xinda Institute of Advanced Technology, 
    \textsuperscript{\rm 6} National University of Singapore\\
    \{yangziqi, wanglijin, yangda, wanjie, zhaoziming, fanzhang, kuiren\}@zju.edu.cn, changec@comp.nus.edu.sg    
}

\maketitle

\begin{abstract}

Neural networks are susceptible to data inference attacks such as the membership inference attack, the adversarial model inversion attack and the attribute inference attack, where the attacker could infer useful information such as the membership, the reconstruction or the sensitive attributes of a data sample from the confidence scores predicted by the target classifier. 
In this paper, we propose a method, namely \codename, to 
defend against membership inference attacks.
It transforms the confidence score vectors predicted by the target classifier and 
makes purified confidence scores indistinguishable in individual shape, 
statistical distribution and prediction label between members and non-members.
The experimental results show that \codename helps defend membership inference attacks with 
high effectiveness and efficiency, outperforming previous defense 
methods, and also incurs negligible utility loss.  
Besides, our further experiments show that \codename is also effective in defending 
adversarial model inversion attacks and attribute inference attacks. For example, the inversion error is raised about 4+ times on the Facescrub530 classifier, and the attribute inference accuracy drops significantly when \codename is deployed in our experiment.

\end{abstract}

\section{Introduction}

Machine learning has been provided as a service by many platforms, transforming various aspects of our daily life such as handling users' sensitive data.
Users access these models through prediction APIs which return a confidence score vector or a label.  
Many studies have indicated that the prediction information of a single sample could be exploited 
to perform data inference attacks to get useful information about this sample on 
which the machine learning model operates
~\cite{shokri_membership_2017, yang_neural_2019,song2020overlearning}.
Data inference attacks could be largely divided into two categories. The first kind of attacks aim at inferring distributional information about a class by observing the prediction changes of different samples~\cite{An.Mirror.NDSS.2022, csmia}, while the second kind of attacks are to infer the individual information of a sample by observing its specific prediction output such as the membership inference attacks~\cite{nasr_machine_2018,salem_ml-leaks_2018,hui_practical_2021,yeom_privacy_2018,li_membership_2020,li_membership_2021}, adversarial model inversion attacks~\cite{yang_neural_2019} and attribute inference attacks~\cite{song2020overlearning}. In this paper, we focus on the second type of data inference attacks.

Among these data inference attacks, \textit{membership inference attack}~\cite{shokri_membership_2017} is one of the most important and exemplary attacks.
In the membership inference attack, the adversary is asked to determine whether a 
given data sample is in the target model's training data.
Many studies acknowledge that the confidence score vectors tell more prediction information beyond the label and thus they should be provided in the prediction results. Therefore, a number of approaches have been proposed to defend the membership inference attack while preserving the confidence scores~\cite{shokri_membership_2017, salem_ml-leaks_2018, nasr_machine_2018, abadi_deep_2016, jia_memguard_2019, selena}.
On the other hand, some studies believe that removing the confidence 
information in the prediction result is a way of defending the membership 
inference attack. However, these defenses are broken by label-only 
attacks~\cite{yeom_privacy_2018, choo_label-only_2020a, li_membership_2021}, 
whereby only the predicted label is exploited to infer the membership.

The major cause of membership inference attack is that the prediction results are distinguishable for members and non-members.  
For example, when a model overfits on the training data, it behaves more confidently on predicting the training data (members) than predicting the testing data (non-members).
The prediction differences between 
members and non-members exist in their \textit{individual shape}, 
\textit{statistical distribution} and \textit{prediction label}.
(1) The target classifier often assigns a higher probability to the predicted class when given a member, making the confidence scores distinguishable in individual shape. This is exploited by many attacks~\cite{salem_ml-leaks_2018, nasr_machine_2018}
(2) Confidence scores on members and non-members are also distinguishable in their
statistical distribution. Our experiments show that confidence scores on the 
members are more clustered in the encoded latent space, while those on  
non-members are more scattered.
BlindMI~\cite{hui_practical_2021} exploits such statistical 
difference to infer membership by comparing the 
distance variation of the confidence scores of two generated datasets.
(3) In addition, the confidence scores on members and non-members are different in the prediction label. Member samples have a higher probability of being correctly classified than the non-member samples, which leads to the difference in classification accuracy. Various label-only attacks exploit such distinguishability~\cite{yeom_privacy_2018, li_membership_2020, li_membership_2021}.

In this paper, we propose a defense method, namely 
\codename, against the membership inference attack.
The main idea is to directly reduce the distinguishability of confidence scores 
and labels on members and non-members by transforming the confidence score 
vectors of members as if they were predicted on non-members. It takes as input the 
prediction produced by the target model and outputs a transformed 
version.  
First, we train \codename on the confidence score vectors predicted by the 
target model on non-member data to reconstruct these vectors using a novel 
training strategy. This encourages \codename to learn the individual shape of 
these non-member confidence score vectors and eventually to generate confidence 
score vectors as if they were drawn from the learned pattern, reducing distinguishability of confidence scores in 
\textit{individual shape}. 
Second, we use Conditional Variational 
Auto-Encoder (CVAE) as a component of \codename to introduce Gaussian noises to the confidence 
scores, such that the statistically clustered confidence scores can be scattered 
and become indistinguishable from those on non-members, reducing distinguishability in \textit{statistical distribution}. 
Third, to decrease the distinguishability in \textit{prediction labels}, 
\codename intentionally modifies the predicted labels of members while preserving those of non-members, which results in a reduction of classification accuracy gap between members and non-members.

Although \codename is designed to defend the membership inference attacks, it turns out to be also 
effective in defending the \textit{adversarial model inversion attack} and the \textit{attribute inference attack}. In the adversarial model inversion attack, the adversary aims at inferring a reconstruction~\cite{yang_neural_2019, fredrikson_model_2015, hitaj_deep_2017} of the data sample.
In the attribute inference attack, the adversary could infer additional sensitive attribute beyond the original input attributes of this data sample~\cite{song2020overlearning}.
We believe that the purification process contributes to the removal of the 
redundant information (hidden in the confidence scores) that is useful to 
recover the input sample, and preserves only the essential semantic information 
for the prediction task. As a result, the adversary can obtain no more useful 
information than the prediction itself from the purified prediction results.

We extensively evaluate \codename on various benchmark datasets and model architectures.
We empirically show that \codename can defend data inference attacks effectively and efficiently with negligible utility loss.
\codename can reduce the membership inference accuracy. For example, the NSH attack~\cite{nasr_machine_2018} accuracy drops from 70.36\% to 51.71\% in our experiments, which is significantly more effective than previous defenses. \codename is also effective against adversarial model inversion attack. For instance, 
the inversion loss on the FaceScrub530 dataset is raised 4+ times (i.e. from 0.0114 to 0.0454) after applying \codename.
Furthermore, \codename can reduce the attribute inference accuracy from 31.06\% to 20.94\% (almost random guessing) on one of evaluated datasets.

\textbf{Contributions.} In summary, we make the following contributions in this paper.
\begin{itemize}
	\setlength\itemsep{0em}
	
	\item To the best of our knowledge, our work is the first to study membership inference attacks comprehensively from the perspectives of \textit{individual shape}, 
	\textit{statistical distribution} and \textit{prediction label}. 
	
	\item We design \codename to 
	defend against membership inference attacks by reducing the 
	distinguishability of the confidence scores in terms of the above three aspects with negligible utility loss. 
	\codename is shown to be also effective in defending other data inference attacks.
 
	\item We extensively evaluate \codename and compare it with existing defenses. Our experimental results show that \codename outperforms existing defenses in both effectiveness and efficiency.

\end{itemize}

\section{Problem Statement}

We focus on classification models of neural 
networks, i.e., a machine learning classifier $F$ is trained on its training dataset $D_{train}$ to map a given sample $\textbf{x}$ to a specific class based on the confidence vectors $F(\vec{x})$ which is the classifier output.  

We consider the data inference attacks designed to infer useful information about a specific sample $\textbf{x}$ based on the target classifier's output $F(\vec{x})$, for examples, the membership inference attack, adversarial model inversion attack and attribute inference attack. We do not consider the data inference attacks~\cite{An.Mirror.NDSS.2022, csmia} which infer distributional information about a class through observing the output changes of $F$ on different $\vec{x}$ in this paper.
\vspace{-0.15cm}
$$
        \vspace{-0.15cm}
	\label{define}
	   \vec{x}, F, D_{aux} \xrightarrow{} \left\{useful\ information\  of\  \vec{x}\right\}
$$

We assume that the attacker has a black-box access to the classifier $F$, where the attacker can only query $F$ with its data sample $\vec{x}$ and obtain the prediction scores $F(\vec{x})$. We also assume that the attacker has an auxiliary dataset $D_{aux}$ to assist its attacks such as a set of data samples drawn from a similar data distribution as the target classifier's training data distribution.

\section{Approach: \codename}

\begin{figure}[t]
        \small
	\begin{center} 
        \vspace{-0.8cm} 
        \setlength{\abovecaptionskip}{-0.5cm}  
        \setlength{\belowcaptionskip}{-0.74cm} 
        \includegraphics[width=1.13\linewidth]{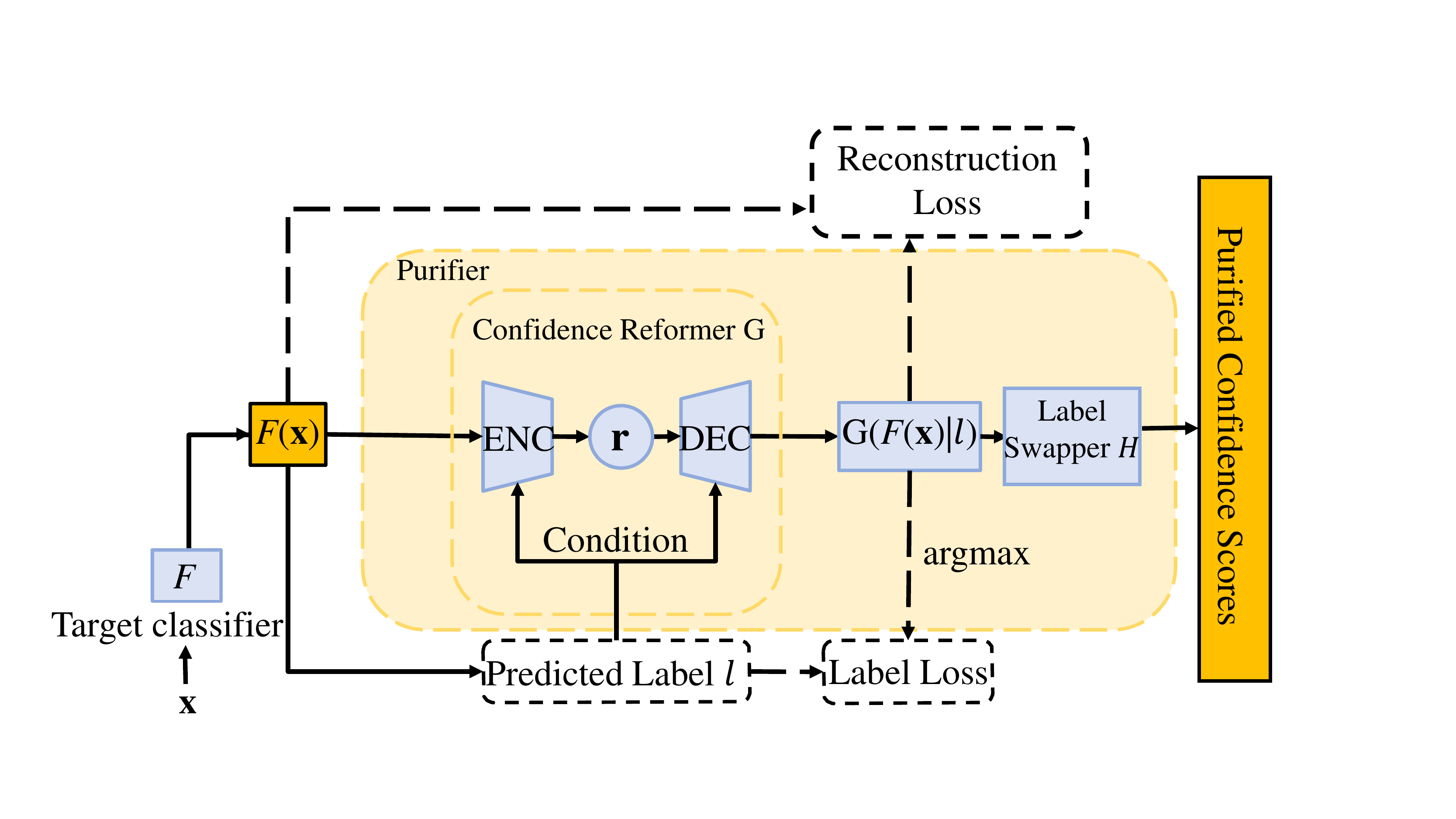}
		\caption{Architecture of \codename. 
           }
		\label{fig:architecture}
	\end{center}
\end{figure}

We propose \codename as a defense of data inference attacks.
Since membership inference attack is one of the most typical instances of the data inference attacks, we design \codename against it as the point of penetration and evaluate its defense performance against other data inference attacks.
The main idea of \codename is to transform the confidence score vectors in such 
a way that they appear indistinguishable on members and non-members. We focus on reducing the three underlying  distinguishabilities of confidence scores between members and non-members: \textit{individual shape}, \textit{statistical distribution} and \textit{prediction label}.

\codename consists of a \textit{confidence reformer} $G$ and a \textit{label swapper} $H$, as shown in Figure~\ref{fig:architecture}. $G$ takes as input the original confidence score vectors
and reforms them as if they were predicted on non-members, achieving indistinguishability of {individual shape} and {statistical distribution}.
We design $G$ as a CVAE, with the predicted label being the condition.
In this way, $G$ is able to learn the overall distribution of the confidence scores from all classes by setting the condition to the corresponding class.
The \textit{label swapper} $H$ takes the reformed confidence scores from $G$, and modifies the predicted labels of members to reduce the gap of classification accuracy between members and non-members, achieving indistinguishability of {prediction label}.

\subsection{Achieving Individual Indistinguishability}

In order to achieve the individual indistinguishability of confidence scores between members and non-members, \codename reforms the confidence scores with the \textit{confidence reformer} $G$, which is a CVAE. $G$ takes the confidence score $F(\vec{x})$ predicted by the target classifier $F$ as input, with the corresponding label $l$ being the condition. $F(\vec{x})$ first goes through the encoder, where it is mapped to the encoded latent space $\vec{r}$. The decoder then maps the confidence score back from the latent space $\vec{r}$,
and the reformed confidence score $G(F(x)|l)$ is obtained. $G$ is trained on the confidence scores predicted by $F$ on the defender’s reference dataset $D_{ref}$, which consists 
of non-member samples. As a result, $G$ learns the pattern of \textit{individual shape} on non-member samples. The reforming process of $G$ could remove any difference in the {individual shape} of $F(\vec{x})$, achieving {individual} indistinguishability.

In order to preserve the classification accuracy, we train $G$
to also produce the label predicted by F by adding a label loss. Formally, $G$ is trained to minimize the following objective function.

\vspace{-0.5cm}
\begin{equation}
	\small
        \vspace{-0.2cm}
	\label{eq:loss}
	\begin{split}
		L(G) = \underset{\vec{x} \sim p_{r}(\vec{x})}{\mathbb{E}} 
		&[\mathcal{R}((G(F(\vec{x})|l), F(\vec{x})) 
		+ \lambda \mathcal{L}((G(F(\vec{x})|l), l)]
	\end{split}
\end{equation}

where $p_r(\vec{x})$ represents the conditional probability of $\vec{x}$ for 
samples in $D_{ref}$, $l$ represents the label of $F(\vec{x})$ (i.e., $l = \argmax(F(\vec{x}))$). $\mathcal{R}$ is a reconstruction loss function ($L_2$ 
norm) and $\mathcal{L}$ is the cross entropy loss function. The parameter $\lambda$ controls the balance of the two loss functions during training.

\subsection{Achieving Statistical Indistinguishability}

We can observe the {statistical distribution} of $F(\vec{x})$ by plotting $F(\vec{x})$ on the encoded latent space $\vec{r}$. Figure~\ref{fig:latent_space_purifier} shows an example of such {statistical distribution} on CIFAR10 dataset, where different 
colors represent different labels. We can observe that confidence score vectors are clustered into several groups according to their labels. However, the members are more densely clustered while non-members are not, which indicates that the distribution of members and non-members are different.

To mitigate the difference in {statistical distribution} between members 
and non-members, \textit{confidence reformer} $G$ introduces Gaussian noises in the latent space $r$, where the label $l$ is used as the condition. During the training process, the reconstruction loss $\mathcal{R}$ encourages the decoder of $G$ to generate confidence scores that are similar to the non-member ones on $D_{ref}$ (non-members) with the same label $l$. However, noises introduced in the latent space $\vec{r}$ will increase the reconstruction error. As a result, $G$ learns a robust latent representation that could preserve the {statistical distribution} of the non-members of label $l$ even if noises are added. During the inference process, the added noises breakdown the clustering of confidence scores on members, while the decoder generates the reformed versions that are similar to the ones on $D_{ref}$, mitigating the difference in {statistical distribution}.

\subsection{Achieving Label Indistinguishability}

To cope with the difference in prediction label, we design a \textit{label swapper} $H$, which modifies the prediction labels of members to reduce the gap of classification accuracy between members and non-members. After training the \textit{confidence reformer} $G$, we randomly select training data to replace their predicted labels with the second largest predicted labels at a certain swap rate $p_{swap} = (acc_{train} - acc_{test} )/ acc_{train}$, where $acc_{train}$ and $acc_{test}$ are the training accuracy and the test accuracy of the target classifier respectively. Note that we fix the data at the training stage whose labels will be modified, so when attackers use the same data to query the final model, they will get the same output. Hence \codename can defend the \textit{replay attack} where attackers exploit the differences between the outputs of multiple same queries to the target model.

Given an input sample $\vec{x}$, $H$ first identifies if $\vec{x}$ is a member whose label needs to be modified.
In order to identify members, the \textit{label swapper} stores information of the original training data. However, it is challenging for the label swapper to efficiently store and index the member information in the run time especially in a large learning task.
To this end,
Label swapper stores $F(\vec{x})$ where $x \in D_{swap}$ as the identifiers to form a prediction indexing set $P_{index}$ whose dimensionality is much smaller than the training data $D_{train}$.
In order to tolerate small perturbations of members added by attackers to indirectly infer membership of a target member sample $x \in D_{swap}$, $H$ uses $k$ nearest neighbor ($k$NN) to identify these suspicious noisy members and also swaps their labels.

\begin{figure}[!t]
        \footnotesize
	\removelatexerror
	\begin{algorithm}[H]
		\SetAlgoLined
		\SetKwInput{KwInput}{Input}                
		\SetKwInput{KwOutput}{Output}             
		\KwInput{The reference dataset $D_{ref}$, the training dataset $D_{train}$, the target classifier $F$, size of mini-batch ${q}$, size of the data need to be modify the labels $t$, number of epochs $P$, learning rate $\eta$, label loss coefficient $\lambda$}
		\KwOut{Model parameters $\theta$ of \textit{label reformer} $G_\theta$, The prediction indexing set $P_{index}$}
		$\theta \gets \mathbf{initialize}(G_\theta)$ \;
		\For{$p=1$ to $P$}{
			\For{each mini-batch ${\{(\vec{x}_{{ref}_{j}},y_{{ref}_{j}})\}}^q_{j=1} \subset D_{ref}$}{
				$c_{r_j} \gets F(\vec{x}_{ref_j})$\;
				$l_{r_j} \gets \mathbf{onehot}(\argmax(c_{r_j})$)\;
				$g \gets \nabla_\theta \frac{1}{q} \sum_{j=1}^{q} \mathcal{R}(G_\theta(c_{r_j}|l_{r_j}),c_{r_j}) + \lambda \mathcal{L}(G_\theta(c_{r_j}|l_{r_j}),l_{r_j})$\;
				$\theta \gets \mathbf{updateParameters}(\eta,\theta,g)$
			}
		}
		$P_{index} \gets \emptyset$\;
            $D_{train} \gets \mathbf{shuffle}(D_{train})$\;
            $D_{swap} \gets {\{(\vec{x}_{{train}_{j}},y_{train_j})\}}^t_{j=1} \subset D_{train}$\;
      		\For{each ${(\vec{x}_{{train}_{j}},y_{train_j})} \in D_{swap}$}{
			$c_j \gets F(\vec{x}_{train_j})$\;
   			$P_{index} \gets P_{index} \cup \{c_j\}$\;
		}
		\Return $G_{\theta}$, $P_{index}$
		\caption{Training process of \codename.}
		\label{alg:train}
	\end{algorithm}
 \vspace{-0.5cm}
\end{figure}

\noindent \underline{\textbf{Training and inference process of \codename}}

The training process of \codename is detailed in Algorithm \ref{alg:train}. For each epoch, we first draw a mini-batch of data points ${\{(\vec{x}_{{ref}_{j}},y_{ref})\}}^q_{j=1}$ from the reference set $D_{ref}$. Then we query the target classifier $F$ to obtain the confidence scores $c_{r_j}$ and the labels $l_{r_j}$ (Line 1-5). After that, the loss is calculated on the objective function~\ref{eq:loss} and gradient descent is used to update the parameters $\theta$ of \textit{confidence reformer} $G$ (Line 6-7). 
When the training of $G$ is finished, we select the data from $D_{train}$ at rate $p_{swap}$ randomly to form $D_{swap}$ (Line 10-11). 
After that, we query the target classifier $F$ to get the confidence $c_{j}$ of the sample ${(\vec{x}_{{train}_{j}},y_{{train}_{j}})} \in D_{swap}$.
The original confidence score $c_{j}$ 
is added to the prediction indexing set $P_{index}$ and later used by the \textit{label swapper} to achieve indistinguishability of prediction label (Line 12-15).

In the inference stage,
given an input sample $\vec{x}$, we first query the target classifier $F$ to get the confidence score $c$ and the predicted label $l$. Then, we input $c$ into the \textit{confidence reformer} $G$, with $l$ being the condition, to get the purified confidence vector $p$.
At this stage, $p$ is indistinguishable in individual shape and statistical distribution.
The \textit{label swapper} $H$ checks if $c$ has a match in $P_{index}$ using $k$NN and swaps the label of $p$ if $c$ is matched. 
This ensures indistinguishability in terms of prediction label. Finally, \codename returns the purified confidence scores $p$.

\section{Experimental Setup}

\subsection{Datasets \& Models}

\noindent \textbf{Membership Inference Attack.}
We use CIFAR10~\cite{shokri_membership_2017, salem_ml-leaks_2018, li_membership_2020}, Purchase100~\cite{shokri_membership_2017, nasr_machine_2018, salem_ml-leaks_2018, li_membership_2020} and 
FaceScrub530~\cite{yang_neural_2019} datasets
which are widely adopted in previous studies on membership 
inference attacks.

\noindent \textbf{Model Inversion Attack}.
We use the same datasets as membership inference attacks. 

\noindent \textbf{Attribute Inference Attack}.
We use the same dataset UTKFace~\cite{zhang2017age} as in a previous study~\cite{song2020overlearning} where the attacker infers additional attribute (i.e., race of five possible values) beyond the original gender classification task.

We attach the details of the datasets to Appendix, including the introduction, pre-processing and data allocation. We also further elucidate the target classifier and \codename on different datasets in the Appendix, including their model architectures and hyper-parameters.

\subsection{Existing Attacks}

In our experiments, we implement the following attacks.

\noindent \textbf{Membership Inference Attack.}
We implement the atttacks including \ding{172} {NSH attack}~\cite{nasr_machine_2018}, \ding{173} {Mlleaks attack}~\cite{salem_ml-leaks_2018}, \ding{174} {Adaptive attack}~\cite{salem_ml-leaks_2018} (where the attacker knows all the details about the defense mechanism), \ding{175} {BlindMI attack}~\cite{hui_practical_2021}, \ding{176} {Label-only attack}~\cite{yeom_privacy_2018, 
li_membership_2020}. \textbf{Model Inversion Attack.}
The attacker trains an inversion model on $D_{aux}$ to perform the model inversion attack. The inversion model takes $F(\vec{x})$ as input and is able to reconstruct $\vec{x}$~\cite{yang_neural_2019}.
\textbf{Attribute Inference Attack.}
The attacker trains a classification on $D_{aux}$ to infer additional sensitive attribute beyond the original input attributes of the given sample~\cite{song2020overlearning}.

We attach the details of the attack above-mentioned methods, including their implementations, to the Appendix.

\subsection{Metrics}

We use the following 4 metrics to measure the utility, defense performance and efficiency of a defense method.

\noindent \ding{172}\textbf{Classification Accuracy:} It is measured on the training set and the test set of the target classifier. 
\ding{173}\textbf{Inference Accuracy:} This is the classification accuracy of 
the attacker's attack model in predicting the membership/sensitive attribute of input samples. 
\ding{174}\textbf{Inversion Error:} Following~\cite{yang_neural_2019}, We measure the inversion error by computing the mean squared error between the original input sample and the reconstruction. 
\ding{175}\textbf{Efficiency:} We measure the efficiency of a defense method by 
reporting its training time and test time relative to the original time required by the target 
classifier. 

\section{Experimental Results}

%\textcolor{red}{1. Defense performance of purifier: }

% \subsection{Defense Performance of \codename}

\begin{table*}[t]
        \scriptsize
  	\setlength{\tabcolsep}{0.4\tabcolsep}
        \setlength{\abovecaptionskip}{-0.01cm}
	\centering
	\caption{Defense performance of \codename against various attacks. Results of Transfer attack and Boundary attack are reported in AUC. Note 
	that the N.A. means that setting is not applicable.         
 }
	\label{tb:bigtable}
	\resizebox{\textwidth}{!}{
		\begin{tabularx}{\linewidth}{c|c|c|c|Y|Y|Y|Y|Y|Y|Y|c}
			\hline
			\multirow{3}{*}{Dataset}      & \multirow{3}{*}{Defense} & 	\multicolumn{2}{c|}{Utility} & \multicolumn{7}{c|}{Membership Inference Attack Accuracy/AUC} & Inversion Error \\ 
						\cline{3-12} 
						
			& & \multirow{2}{*}{Train acc} &  \multirow{2}{*}{Test acc.}  &  \multirow{2}{*}{NSH} &  \multirow{2}{*}{Mlleaks} &  \multirow{2}{*}{Adaptive} & \multirow{2}{*}{BlindMI} & \multicolumn{3}{c|}{Label only attacks}  & \multirow{2}{*}{$L_2$ norm}         \\ 
			\cline{9-11}
			& &  &  & &  &  &  & Gap & Transfer  & Boundary  &        \\ 
			\hline
			\multirow{2}{*}{CIFAR10}      & None                     & 
			99.99\%       & 95.92\%      & 56.03\% & 56.26\% & N.A.      & 
			54.76\% & 52.04\%    & 0.5048              & 0.5214              & 
			1.4357          \\ \cline{2-12} 
			& Purifier                 & 97.60\%       & 95.92\%      & 51.65\% 
			& 50.26\% & 50.23\%   & 50.64\% & 50.84\%    &0.4974              
			& 0.4949              & 1.4939          \\ \hline
			\multirow{2}{*}{Purchase100}  & None                     & 
			100.00\%      & 84.36\%      & 70.36\% & 64.43\% & N.A.      & 
			69.82\%    & 57.82\%    & 0.5431              & N.A.                & 
			0.1426          \\ \cline{2-12} 
			& Purifier                 & 86.59\%       & 83.23\%      & 51.71\% 
			& 50.09\% & 50.13\%   & 50.96\%
			    & 51.68\%    &0.4978             
			& N.A.                & 0.1520          \\ \hline
			\multirow{2}{*}{FaceScrub530} & None                     & 
			100.00\%      & 77.68\%      & 69.34\% & 75.04\% & N.A.      & 
			62.61\% & 61.16\%    & 0.5869              & 0.7739              & 
			0.0114          \\ \cline{2-12} 
			& Purifier                 & 77.58\%       & 77.52\%      & 51.56\% 
			& 51.04\% & 50.00\%   & 50.00\% & 50.02\%    &  0.4983             
			&0.6185              & 0.0454          \\ \hline
		\end{tabularx}
	}
        \vspace{-0.5cm}
\end{table*}

\subsection{\codename is Effective in Membership Inference}

\noindent \underline{\textbf{Effectiveness}}

Table~\ref{tb:bigtable} presents the  defense 
performance of \codename against different membership inference attacks. 
For each classification task, \codename decreases the attack 
accuracy as well as preserves the the classification accuracy.
\codename reduces the accuracy of NSH attack significantly for 
different datasets. For instance, it reduces the accuracy of NSH attack 
from 69.34\% to 51.56\% in FaceScrub530 dataset. 
As for Mlleaks attack, the model defended with \codename reduces 
the attack accuracy to nearly 50\%.
Comparing with the pure Mlleaks attack, the performance of the adaptive attack does not show a large difference where \codename reduces the accuracy to nearly 50\%.
\codename is also effective against BlindMI attack. For example, \codename reduces the accuracy of BlindMI from 62.61\% to 50.00\% in FaceScrub530 
dataset.

\noindent \underline{\textbf{Comparison with other defenses}}

\noindent \textbf{Existing Defenses.}
We compare \codename with following defenses.
\ding{172}Min-Max~\cite{nasr_machine_2018}.
\ding{173}MemGuard~\cite{jia_memguard_2019}.
\ding{174}Model-Stacking~\cite{salem_ml-leaks_2018}.
\ding{175}MMD Defense~\cite{li_membership_2021}. 
\ding{176}SELENA~\cite{selena}.
\ding{177}One-Hot Encoding. 
\ding{178}Random Noise.
We attach the details of the above-mentioned defense methods, including their implementations and selections of hyperparameters, to the Appendix.

\begin{table*}[t]
	\tiny
        \vspace{0.2cm}
        \setlength{\abovecaptionskip}{-0.01cm}
	\centering
	\caption{Defense performance of \codename and other defense methods. }
	\label{tb:comparetable}
	\resizebox{\textwidth}{!}{
		\begin{tabular}{c|c|c|c|c|c|c|c|c}
			\hline
			{\color[HTML]{000000} Dataset}                        & 
			{\color[HTML]{000000} Defense}          & {\color[HTML]{000000} 
			Training acc.} & {\color[HTML]{000000} Test acc.} & 
			{\color[HTML]{000000} NSH Attack}       & {\color[HTML]{000000} 
			Mlleaks Attack} &BlindMI Attack  &Gap Attack  & Inversion Error                        \\ \hline
			{\color[HTML]{000000} }                               & 
			{\color[HTML]{000000} Purifier}         & {\color[HTML]{000000} 
			97.60\%}       & {\color[HTML]{000000} 95.92\%}   & 
			{\color[HTML]{000000} \textbf{51.65\%}} & {\color[HTML]{000000} 
			50.26\%}     &  \textbf{50.64\%} & \textbf{50.84\%}  & {\color[HTML]{000000} \textbf{1.4939}} \\
			{\color[HTML]{000000} }                               & 
			{\color[HTML]{000000} Min-Max}          & {\color[HTML]{000000} 
			99.40\%}       & {\color[HTML]{000000} 94.38\%}   & 
			{\color[HTML]{000000} 53.97\%}          & {\color[HTML]{000000} 
			52.93\%}   & 53.52\%  &   52.51\%  & {\color[HTML]{000000} 1.4770}          \\
			{\color[HTML]{000000} }                               & 
			{\color[HTML]{000000} MemGuard}         & {\color[HTML]{000000} 
			99.99\%}       & {\color[HTML]{000000} 95.92\%}   & 
			{\color[HTML]{000000} 53.63\%}          & {\color[HTML]{000000} 
			52.24\%}     &  52.03\% &  52.04\% & {\color[HTML]{000000} 1.4439}          \\
			{\color[HTML]{000000} }                               & 
			{\color[HTML]{000000} Model-Stacking}   & {\color[HTML]{000000} 
			95.80\%}       & {\color[HTML]{000000} 92.12\%}   & 
			{\color[HTML]{000000} 51.93\%}          & {\color[HTML]{000000} 
			51.01\%}  &  52.69\%  &  51.84\%  & {\color[HTML]{000000} 1.4723}          \\
			{\color[HTML]{000000} }                               & 
			{\color[HTML]{000000} MMD Defense}      & {\color[HTML]{000000} 
			99.99\%}       & {\color[HTML]{000000} 87.44\%}   & 
			{\color[HTML]{000000} 59.50\%}          & {\color[HTML]{000000} 
			57.60\%}     & 58.92\% &  56.28\% & {\color[HTML]{000000} 1.4414}          \\
			{\color[HTML]{000000} }                               &
   {\color[HTML]{000000} SELENA}     & {\color[HTML]{000000} 
			98.40\%}       & {\color[HTML]{000000} 93.90\%}   & 
			{\color[HTML]{000000} 52.14\%}          & {\color[HTML]{000000} 
			{52.35\%}} & 51.08\%	&	52.25\% & {\color[HTML]{000000} 1.4350}      
                \\
                {\color[HTML]{000000} }           & 
                 {\color[HTML]{000000} One-Hot Encoding} & {\color[HTML]{000000} 
			99.99\%}       & {\color[HTML]{000000} 95.92\%}   & 
			{\color[HTML]{000000} 52.17\%}          & {\color[HTML]{000000} 
			\textbf{50.00\%}}      &    51.88\%  & 52.04\% & {\color[HTML]{000000} 1.4414} 
			           \\ 
                \multirow{-8}{*}{{\color[HTML]{000000} CIFAR10}}      &
                   {\color[HTML]{000000} Random Noise}     & {\color[HTML]{000000} 
			99.99\%}       & {\color[HTML]{000000} 95.92\%}   & 
			{\color[HTML]{000000} 55.97\%}          & {\color[HTML]{000000} 
			{50.01\%}} & 51.69\%	&	52.04\% & {\color[HTML]{000000} 1.4342}         \\ \hline
			{\color[HTML]{000000} }                               & 
			{\color[HTML]{000000} Purifier}         & {\color[HTML]{000000} 
			86.59\%}       & {\color[HTML]{000000} 83.23\%}   & 
			{\color[HTML]{000000} \textbf{51.71\%}} & {\color[HTML]{000000} 
			{50.09\%}} & \textbf{50.96\%}  &\textbf{51.68\%} & {\color[HTML]{000000} {0.1520}} \\
			{\color[HTML]{000000} }                               & 
			{\color[HTML]{000000} Min-Max}          & {\color[HTML]{000000} 
			99.89\%}       & {\color[HTML]{000000} 82.03\%}   & 
			{\color[HTML]{000000} 65.13\%}          & {\color[HTML]{000000} 
			63.95\%}     &   57.39\%  & 58.93\% & {\color[HTML]{000000} 0.1428}          \\
			{\color[HTML]{000000} }                               & 
			{\color[HTML]{000000} MemGuard}         & {\color[HTML]{000000} 
			100.00\%}      & {\color[HTML]{000000} 84.36\%}   & 
			{\color[HTML]{000000} 62.28\%}          & {\color[HTML]{000000} 
			57.86\%}      &  61.35\%  & 57.82\% & {\color[HTML]{000000} 0.1426}          \\
			{\color[HTML]{000000} }                               & 
			{\color[HTML]{000000} Model-Stacking}   & {\color[HTML]{000000} 
			81.84\%}       & {\color[HTML]{000000} 69.68\%}   & 
			{\color[HTML]{000000} 61.16\%}          & {\color[HTML]{000000} 
			55.53\%}   &  60.36\%   &56.08\%  & {\color[HTML]{000000} 0.1472}          \\
			{\color[HTML]{000000} }                               & 
			{\color[HTML]{000000} MMD Defense}   & {\color[HTML]{000000} 100.00\%}       & {\color[HTML]{000000} 82.65\%}   & 
{\color[HTML]{000000} 69.48\%}          & {\color[HTML]{000000} 
69.89\%}     &  66.62\% & 58.67\% & {\color[HTML]{000000} 0.1439}          \\		
			{\color[HTML]{000000} }                               &
         			{\color[HTML]{000000} SELENA}     & {\color[HTML]{000000} 
			83.24\%}       & {\color[HTML]{000000} 79.53\%}   & 
                {\color[HTML]{000000} 51.90\%}          & {\color[HTML]{000000} 
			52.97\%}    &   53.04\% & 51.83\% & {\color[HTML]{000000} 0.1440} 
			           \\
			\multirow{-6}{*}{{\color[HTML]{000000} Purchase100}}  &
   {\color[HTML]{000000} One-Hot Encoding} & {\color[HTML]{000000} 
			100.00\%}      & {\color[HTML]{000000} 84.36\%}   & 
			{\color[HTML]{000000} 57.65\%}          & {\color[HTML]{000000} 
			\textbf{50.00\%}}      &   57.67\% & 57.82\% & \textbf{\color[HTML]{000000} 0.1524}
     \\ 
                {\color[HTML]{000000} } &
                {\color[HTML]{000000} Random Noise}     & {\color[HTML]{000000} 
			100.00\%}       & {\color[HTML]{000000} 84.36\%}   & 
			{\color[HTML]{000000} 60.06\%}          & {\color[HTML]{000000} 
			50.02\%}    &   54.44\% & 57.82\% & {\color[HTML]{000000} 0.1409}
        \\
                \hline
			{\color[HTML]{000000} }                               & 
			{\color[HTML]{000000} Purifier}         & {\color[HTML]{000000} 
			77.58\%}       & {\color[HTML]{000000} 77.52\%}   & 
			{\color[HTML]{000000} \textbf{51.56\%}} & {\color[HTML]{000000} 
			{51.04\%}} & \textbf{50.00\%} & \textbf{50.03\%} &{\color[HTML]{000000} \textbf{0.0454}} \\
			{\color[HTML]{000000} }                               & 
			{\color[HTML]{000000} Min-Max}          & {\color[HTML]{000000} 
			98.99\%}       & {\color[HTML]{000000} 68.31\%}   & 
			{\color[HTML]{000000} 65.56\%}          & {\color[HTML]{000000} 
			69.84\%}     & 66.16\%& 65.34\% & {\color[HTML]{000000} 0.0182}          \\
			{\color[HTML]{000000} }                               & 
			{\color[HTML]{000000} MemGuard}         & {\color[HTML]{000000} 
			100.00\%}      & {\color[HTML]{000000} 77.68\%}   & 
			{\color[HTML]{000000} 62.48\%}          & {\color[HTML]{000000} 
			60.06\%}     &  62.72\% & 61.16\% & {\color[HTML]{000000} 0.0117}          \\
			{\color[HTML]{000000} }                               & 
			{\color[HTML]{000000} Model-Stacking}   & {\color[HTML]{000000} 
			86.30\%}       & {\color[HTML]{000000} 57.05\%}   & 
			{\color[HTML]{000000} 62.00\%}          & {\color[HTML]{000000} 
			51.86\%}       & 60.62\% & 64.63\% & {\color[HTML]{000000} 0.0417}          \\
			{\color[HTML]{000000} }                               & 
			{\color[HTML]{000000} MMD Defense}      & {\color[HTML]{000000} 
			100.00\%}      & {\color[HTML]{000000} 77.38\%}   & 
			{\color[HTML]{000000} 64.88\%}          & {\color[HTML]{000000} 
			67.95\%}    &  63.55\% &  61.31\% & {\color[HTML]{000000} 0.0111}          \\
			{\color[HTML]{000000} }                               &
      {\color[HTML]{000000} SELENA}     & {\color[HTML]{000000} 
			81.06\%}       & {\color[HTML]{000000} 72.05\%}   & 
			{\color[HTML]{000000} 51.68\%}          & {\color[HTML]{000000} 
			51.23\%}  &  54.05\%  &  50.50\%  & {\color[HTML]{000000} 0.0131}
			           \\
			\multirow{-7}{*}{{\color[HTML]{000000} FaceScrub530}} &
   {\color[HTML]{000000} One-Hot Encoding} & {\color[HTML]{000000} 
			100.00\%}      & {\color[HTML]{000000} 77.68\%}   & 
			{\color[HTML]{000000} 57.87\%}          & {\color[HTML]{000000} 
			\textbf{50.00\%}}        &   61.23\%  & 61.16\% &{\color[HTML]{000000} 0.0420}
			           \\ {\color[HTML]{000000} } &
              {\color[HTML]{000000} Random Noise}     & {\color[HTML]{000000} 
			100.00\%}       & {\color[HTML]{000000} 77.68\%}   & 
			{\color[HTML]{000000} 56.85\%}          & {\color[HTML]{000000} 
			50.04\%}  &  60.83\%  &  61.16\%  & {\color[HTML]{000000} 0.0175}
                       \\ 
            \hline
		\end{tabular}
	}
        \vspace{-0.5cm}
\end{table*}

Table~\ref{tb:comparetable} shows the defense performance of \codename and 
other defense methods against membership inference attacks under different 
datasets. 
\codename achieves the best defense 
performance against most of the attacks, including the NSH attack, the BlindMI attack and the gap attack compared to other 
methods on all of three datasets. For the Mlleaks Attack, \codename can achieve the second best performance only to One-Hot Encoding and Random Noise. 
\codename also achieves a better security-utility 
tradeoff than other defenses. It imposes a reduction in test 
accuracy of about 1\%. In comparison, Model-Stacking and SELENA can mitigate membership inference attacks to some 
extent, but they incur intolerable reduction in model's test accuracy.
For One-Hot 
Encoding and Random Noise,
their transformation on 
confidence vectors leads to a large degree of semantic information loss.
MemGuard reaches acceptable defense performance with negligible decline in test 
accuracy. However, its defense performance is not as good as that of \codename.

\subsection{\codename is Effective in Adversarial Model Inversion}

\noindent \underline{\textbf{Effectiveness}}

\begin{figure}[t]
	\centering 
        \vspace{-0.15cm}
        \setlength{\abovecaptionskip}{-0.1cm}
        \setlength{\belowcaptionskip}{-0.4cm} 
	\subfigure{
		\includegraphics[width=0.85\linewidth]{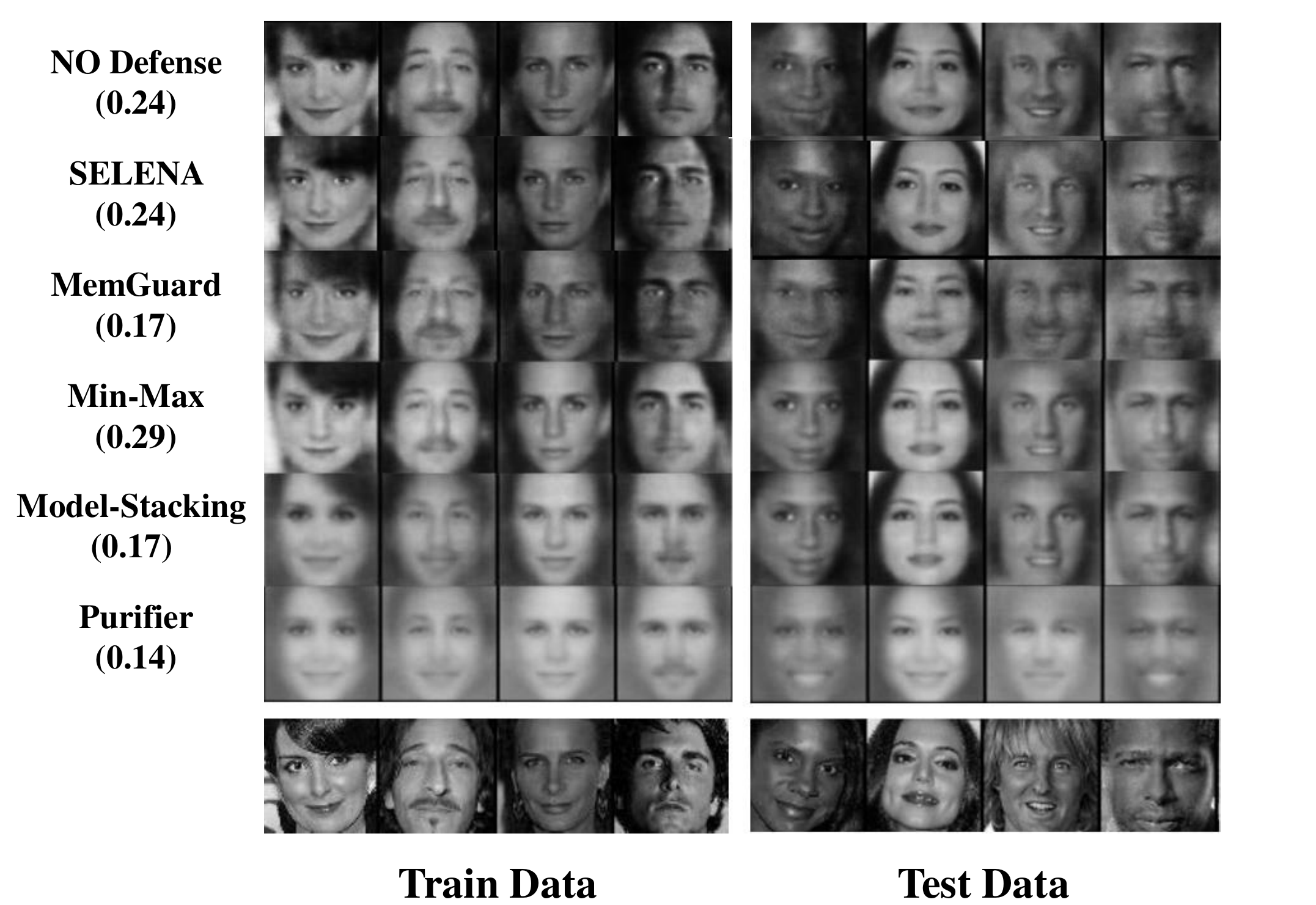}
	}
	\caption{Model inversion attack against the FaceScrub530 classifier 
		defended by different approaches. 
  }
	\label{fig:facescrub}
        \vspace{-0.3cm}
\end{figure}

We further investigate the defense performance of \codename against adversarial model 
inversion attack.
We train an inversion attack model on top of each classifier 
with or without defense on FaceScrub530 dataset.
Although \codename is designed to protect models from membership 
inference attacks, it turns out that the \codename is also effective in mitigating model inversion attack. Figure~\ref{fig:facescrub} 
shows the results of our 
experiment on adversarial model inversion attack on FaceScrub530.
We quantify the inversion quality by reporting the average facial similarity scores compared with the ground truth using the Microsoft Azure Face Recognition service~\cite{azure_notitle_2020}, which is shown on left side of Figure~\ref{fig:facescrub}. The less the number is,the less similarity reconstructed samples share with the original samples.  

We report all the inversion error under three datasets in Table~\ref{tb:bigtable}. As shown in Table~\ref{tb:bigtable} and Figure~\ref{fig:facescrub}, the inversion loss on the FaceScrub530 dataset is raised 4+ times (i.e. from 0.0114 to 0.0454) after applying \codename, indicating the performance reduction of the inversion attack is significant. 
Note that the effect of defense against the adversarial model inversion attacks on Purchase100 and CIFAR10 seems less significant compared with FaceScrub530
This is because the inversion attack does not perform well on these classifiers even though without any defense.

\noindent \underline{\textbf{Comparison with other defenses}}

\codename also achieves the best performance in defending model inversion attack on CIFAR10 and Facescrub530. 
Table~\ref{tb:comparetable} shows that \codename has the largest inversion 
error(also called reconstruction error) compared with other defenses on these datasets, 
quantitatively demonstrating that \codename achieves better defense performance 
against adversarial model inversion attack than other defenses. 
Figure~\ref{fig:facescrub} depicts the reconstructed samples from confidence 
vectors given by each defense model on FaceScrub530 dataset. With \codename as 
defense, the reconstructed images are much less similar to the ground truth 
image and look more blurred. Other defense methods, however, could not protect 
the model from adversaries recovering small details of the original image. It 
can be quantitatively verified by the similarity scores gathered from the 
Microsoft Azure Face Recognition service. For instance, the average similarity scores 
of reconstructed images of MemGuard-defended models are 0.17, which are larger 
than that of \codename (i.e., 0.14).
\codename achieves the smallest similarity scores among other defense 
methods, indicating that \codename can protect the target model against 
adversarial model inversion attack effectively.

\subsection{\codename is Effective in Attribute Inference}

\noindent \underline{\textbf{Effectiveness}}

We deploy \codename under the attribute inference attack and find that \codename is also effective in mitigating it. We train an attribute inference classifier on UTKFace dataset to predict the race of the given sample. 
Table~\ref{tb:attributeinference} shows the results of our experiment.
The attribute inference accuracy on the UTKFace dataset is reduced to 20.94\% (almost random guessing) after applying \codename.

\begin{table}[t]
	\centering
        \vspace{-0.4cm}
        \scriptsize
        \vspace{0.2cm}
        \setlength{\abovecaptionskip}{-0.01cm} 
        \setlength{\belowcaptionskip}{0.6cm}
	\caption{Attribute inference attack against the UTKFace
classifier with and without \codename.}
	\label{tb:attributeinference}
	\resizebox{\columnwidth}{!}{
		\begin{tabularx}{\linewidth}{c|Y|Y|Y|c}
			\hline
			\multirow{2}{*}{Dataset}      & \multirow{2}{*}{Defense} & 	\multicolumn{2}{c|}{Utility} & \multirow{2}{*}{Attack Accuracy} \\ 
						\cline{3-4} 
						
			& & \multirow{1}{*}{Train acc} &  \multirow{1}{*}{Test acc.} \\         
			\hline 
			\multirow{2}{*}{UTKFace}      & None                     & 
			99.92\%       & 83.08\%      & 31.06\%   \\ 
                \cline{2-5} 
			& Purifier    & 84.20\%       & 82.78\%      & 20.94\%     \\ \hline
 
		\end{tabularx}
	}
        \vspace{-0.65cm}
\end{table}

\begin{figure}[t]
        \vspace{0.3cm}
        \setlength{\abovecaptionskip}{-0.2cm}
        \setlength{\belowcaptionskip}{-0.5cm}
	\centering  
        \tiny
	\begin{minipage}[b]{1\linewidth}
		\centering
		
		\subfigure{
			\includegraphics[width=0.5\linewidth]{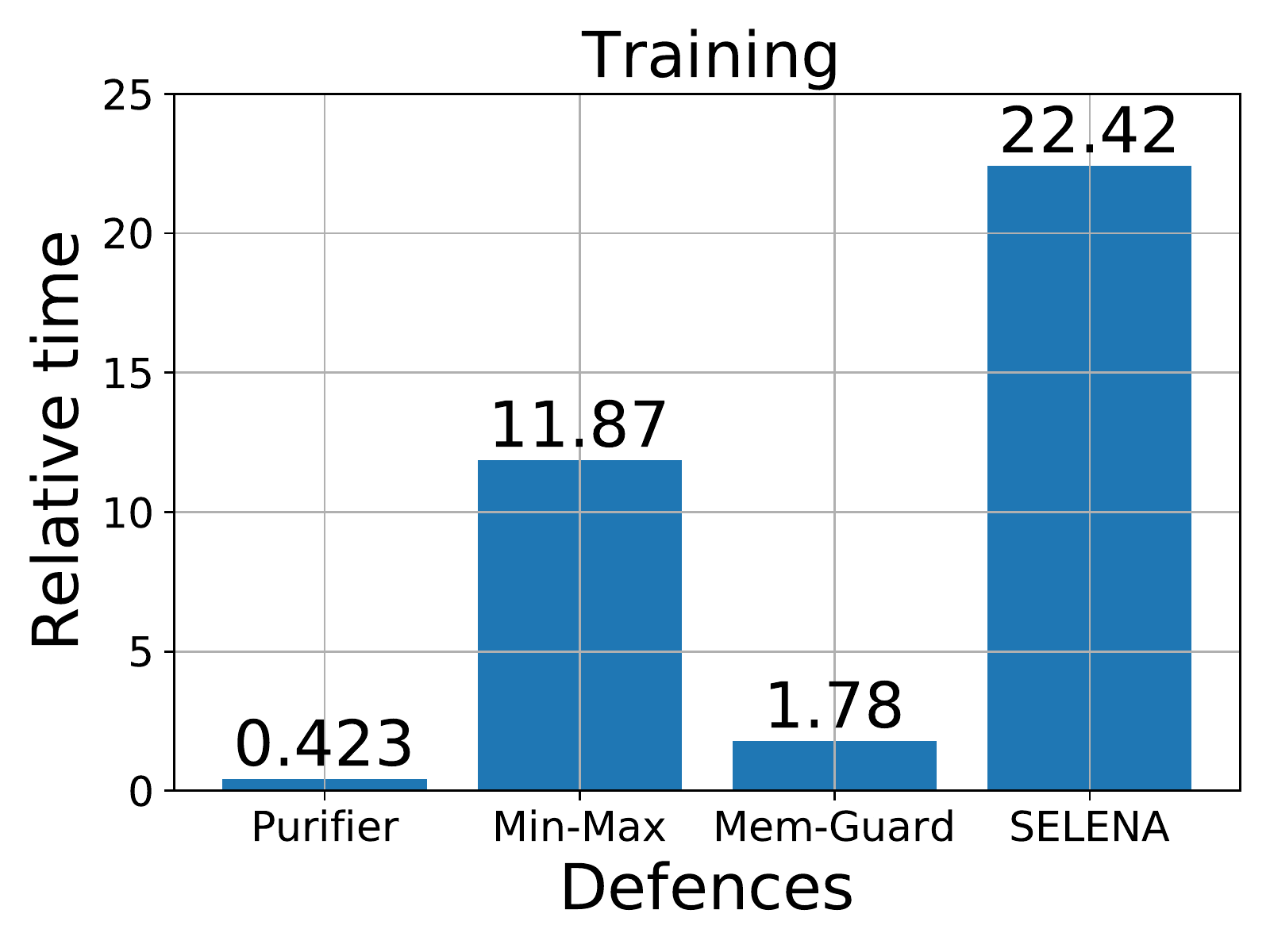}
		\includegraphics[width=0.5\linewidth]{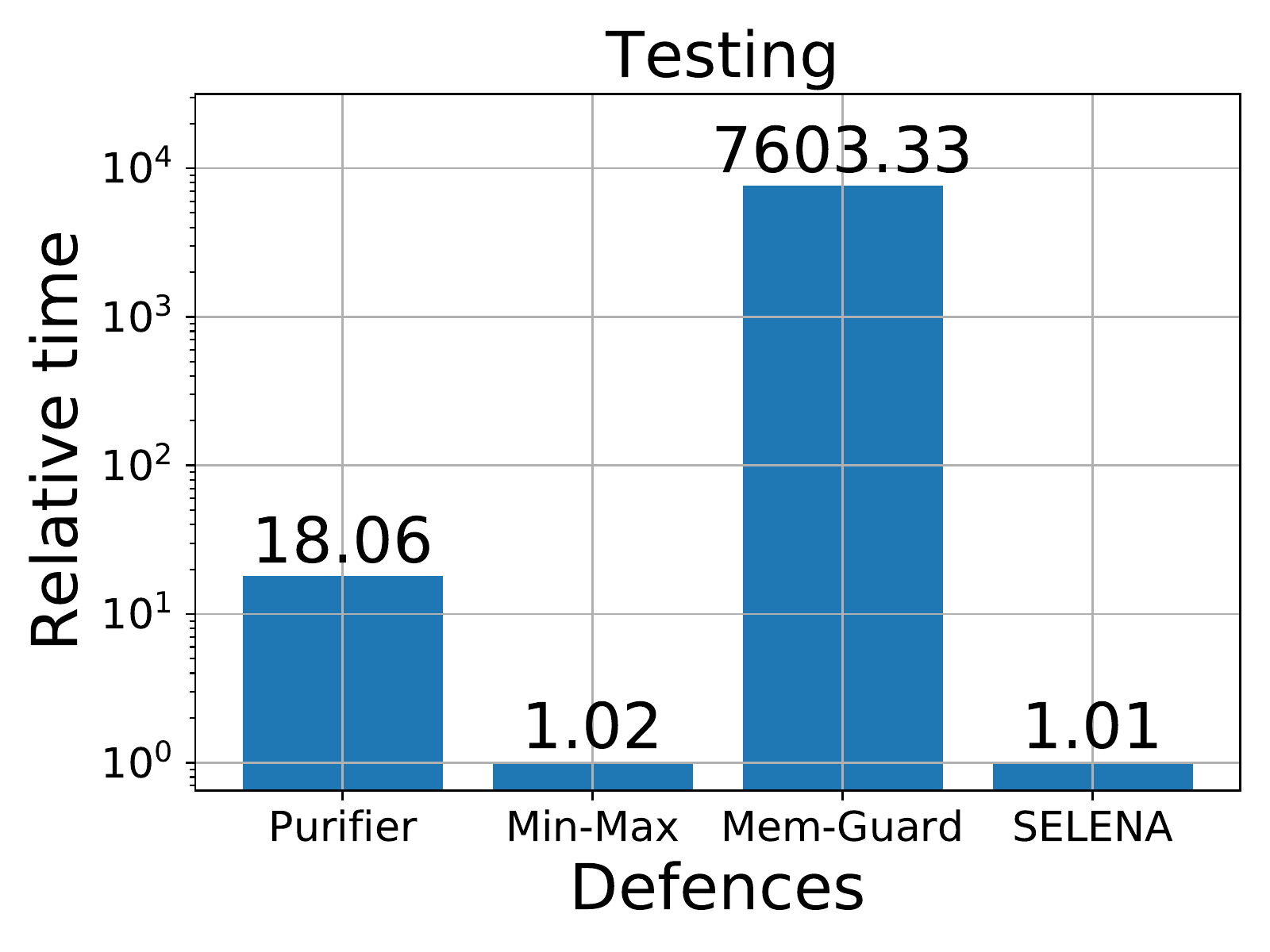}
		}		
	\end{minipage}
	\caption{Efficiency of different defense methods. 
  }
	\label{fig:efficiency}
  \vspace{-0.2cm}
\end{figure}

\subsection{Efficiency} 
Figure~\ref{fig:efficiency} presents the efficiency of \codename compared with other defenses. 
We perform our experiments on a PC equipped with four Titan XP GPUs with 12GBytes of graphic memory,128 GBytes of memory and an Intel Xeon E5-2678 CPU.
The training time of \codename is only 0.423 time of the target classifier, which outperforms all the other methods. 
The testing time of \codename is {18.06} times as much as the target
classifier, which is considered acceptable compared to MemGuard whose testing time is 7,000+ times more than the original classifier.

\subsection{Analysis of Purified Confidence Scores}

In this subsection, we analyze how the purified confidence scores affect membership inference attacks by evaluating three indistinguishabilities: individual, statistical and label.

\noindent \underline{\textbf{Individual Indistinguishability of Purified 
		Confidence}}

\begin{figure}[t]
	\centering
 	\subfigbottomskip=1pt
        \setlength{\abovecaptionskip}{0.2cm}
        \setlength{\belowcaptionskip}{-0.5cm}
	\begin{minipage}[b]{1\linewidth}
		\centering
		\subfigure{
			\includegraphics[width=0.5\linewidth]{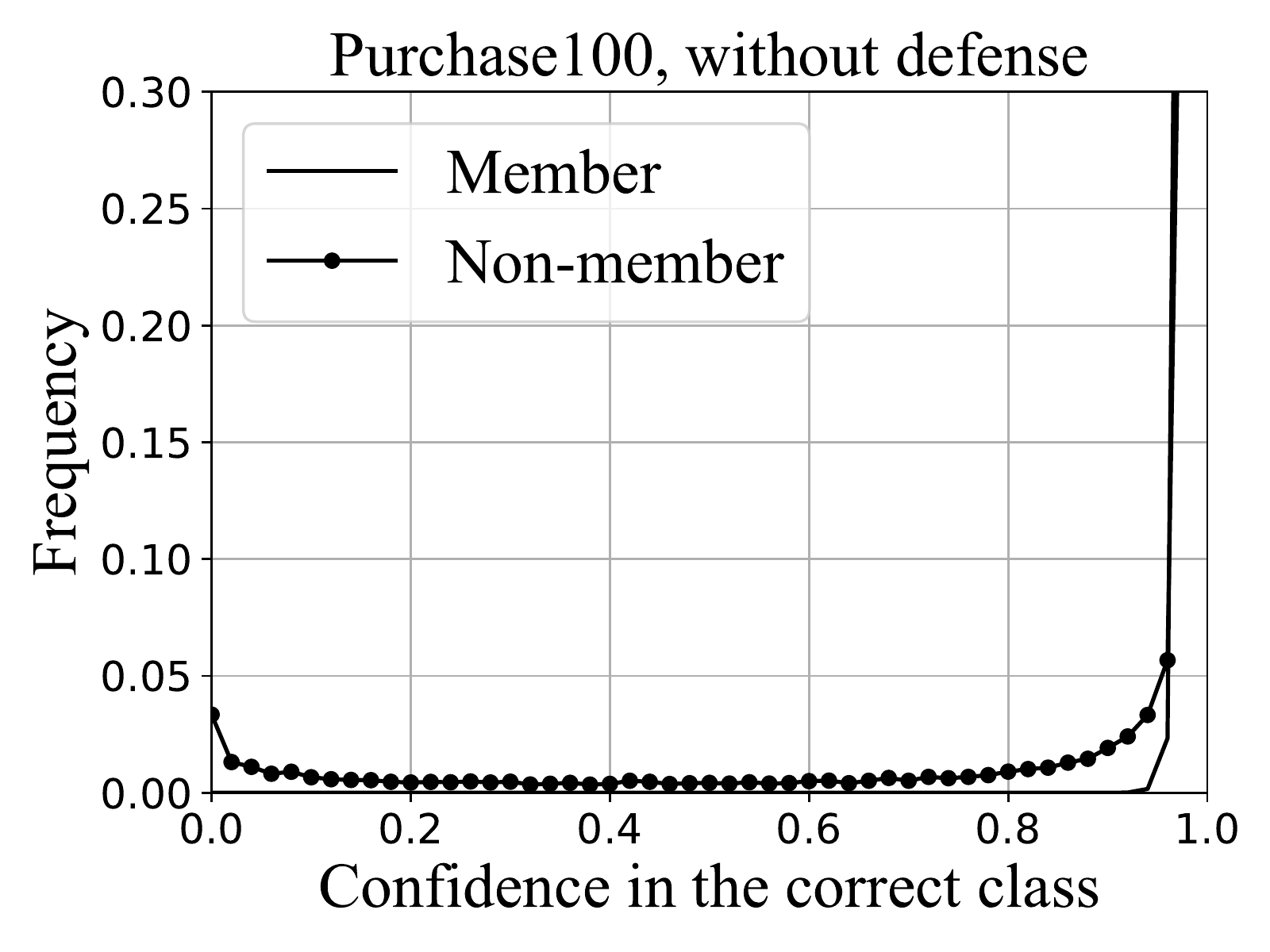}
			\includegraphics[width=0.5\linewidth]{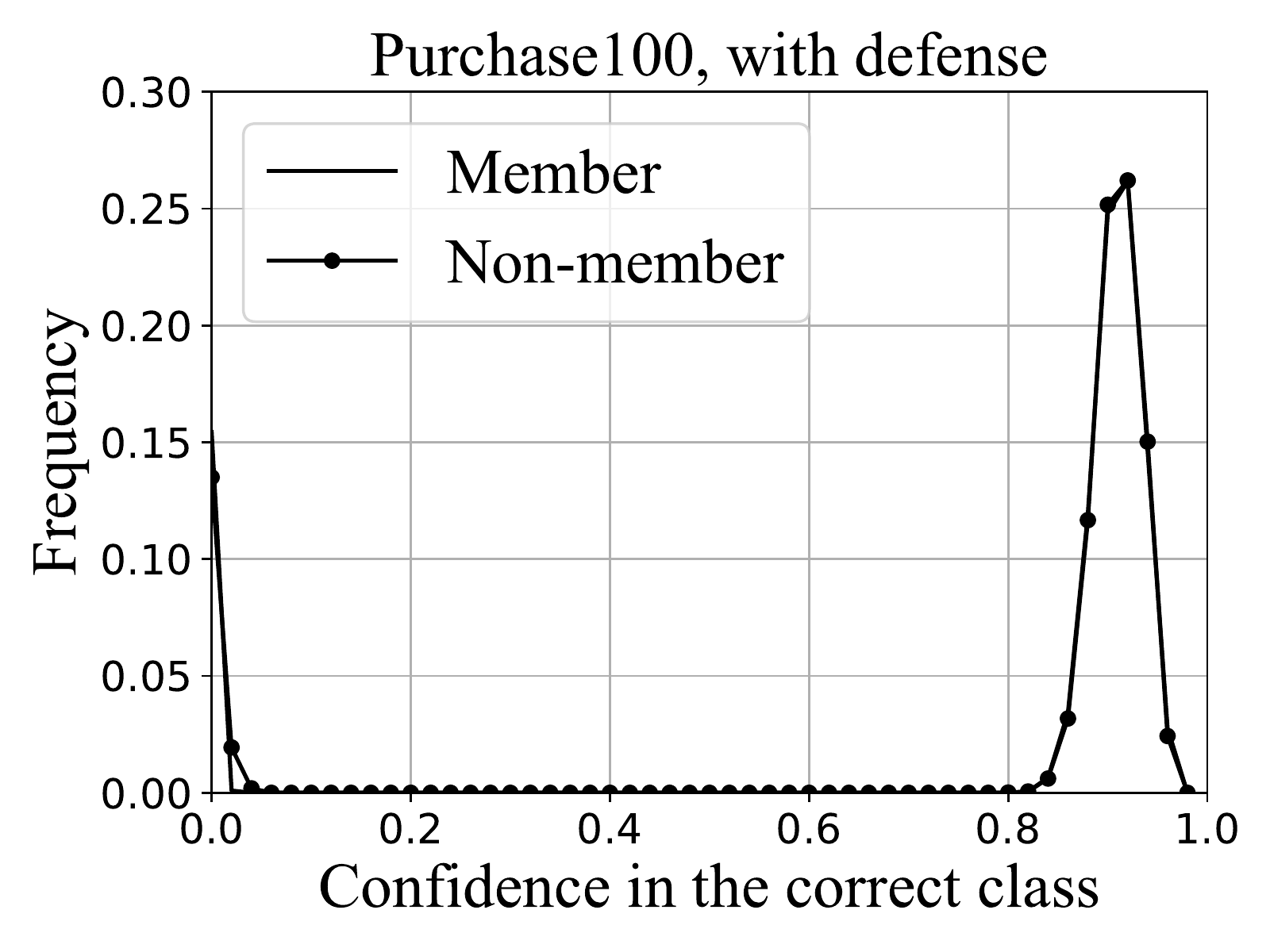}
		}	
        \vspace{-0.1cm}
	\end{minipage}
    \vspace{-0.3cm} 
 \begin{minipage}[b]{1\linewidth}
		\centering
            \setlength{\abovecaptionskip}{-0.2cm}
		 
		\subfigure{
			\includegraphics[width=0.5\linewidth]{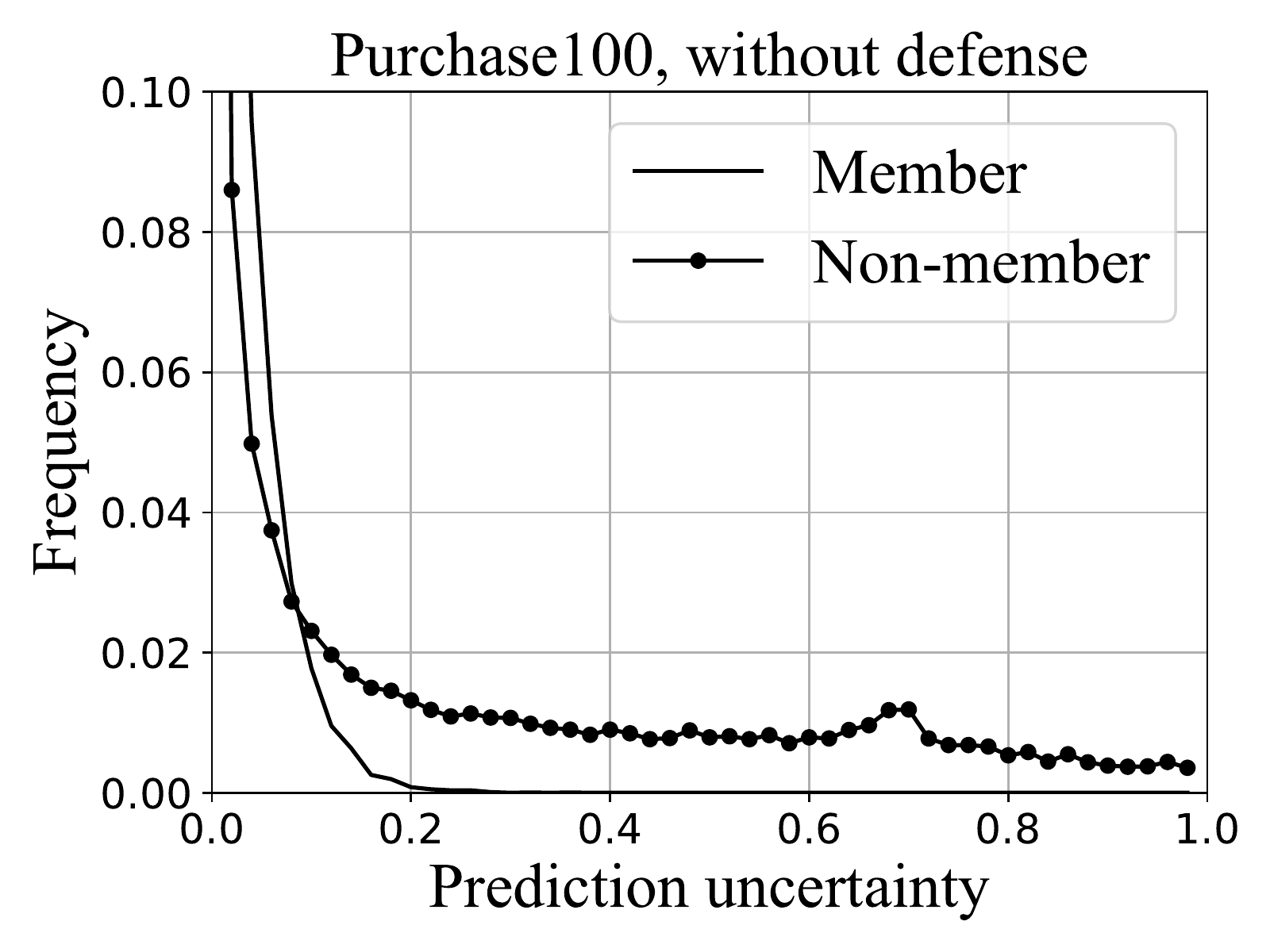}
			\includegraphics[width=0.5\linewidth]{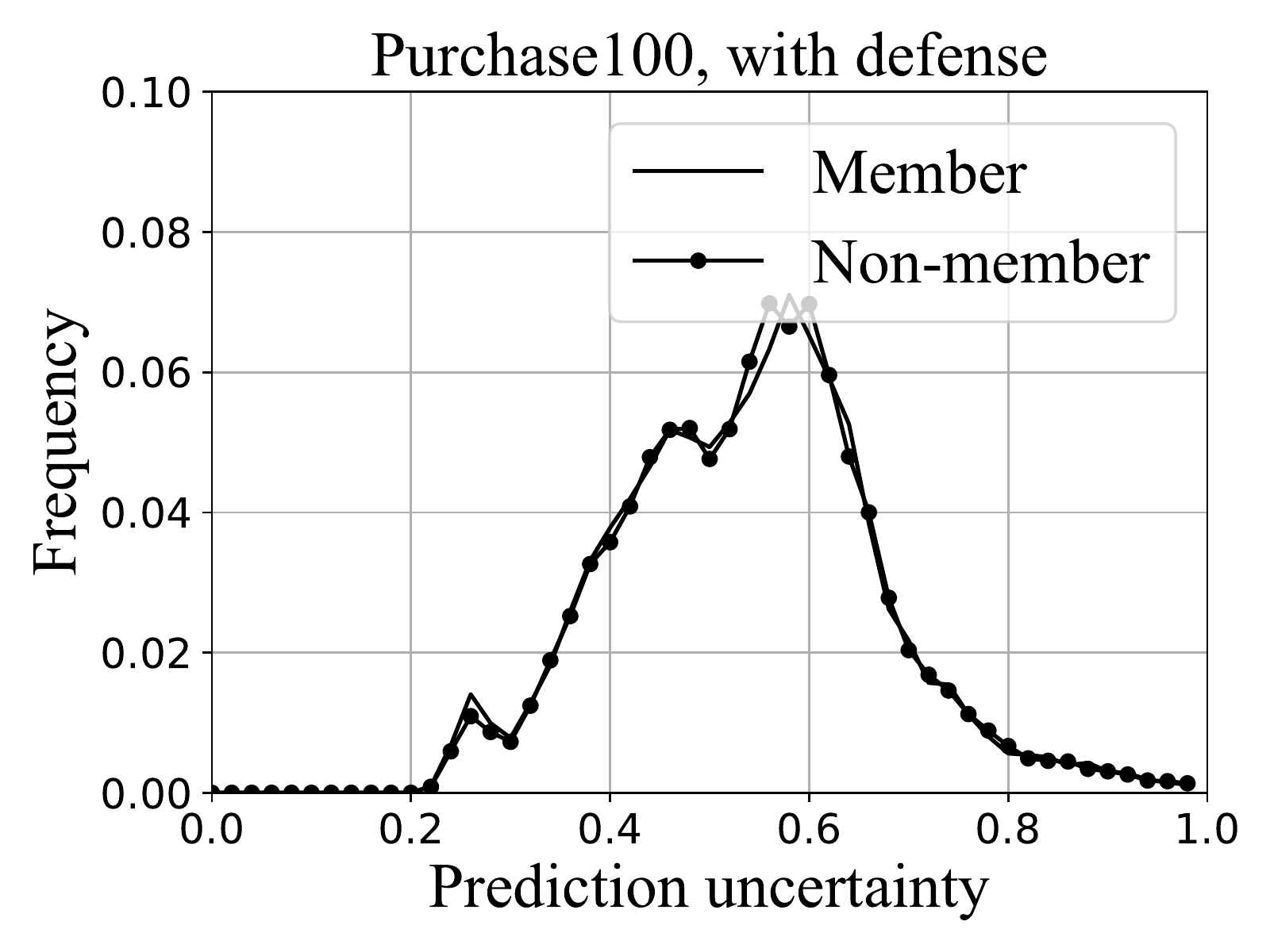}
		}
		
	\end{minipage}
	
	\caption{Distribution of the target classifier's confidence in predicting the correct class and the prediction uncertainty on members and non-members of training set.}
	\label{fig:prediction_acc}
        \vspace{-0.2cm}
\end{figure}

\newcolumntype{P}[1]{>{\Centering\hspace{0pt}}p{#1}}
\newcolumntype{Z}{>{\centering\let\newline\\\arraybackslash\hspace{0pt}}X}

\begin{table}[t]
	\centering
	\scriptsize
        \setlength{\abovecaptionskip}{-0.02cm}
        \setlength{\belowcaptionskip}{0.5cm}
	\caption{Gap of the classifier's confidence in predicting the correct class(i.e, Confi) 
		and the prediction uncertainty(i.e, Uncer) between members and non-members.}
	\label{tb:measurement}
	\resizebox{\columnwidth}{!}{
		\begin{tabularx}{\columnwidth}{l|l|Y|Y|Y|Y|Y|Y}
			\hline
			\multirow{2}{*}{Metric} & \multirow{2}{*}{Defense} & 
			\multicolumn{2}{Y|}{CIFAR10} & \multicolumn{2}{Y|}{Purchase100} & 
			\multicolumn{2}{Y}{FaceScrub530} \\
			\hhline{~|~|-|-|-|-|-|-}
			&  & Max & Avg. & Max & Avg. & Max & Avg. \\
			\hline
			\multirow{2}{*}{Confi} & None & 0.103 & 0.004 & 0.412 & 0.016 & 
			0.415 & 0.017 \\
			\cline{2-8}
			& Purifier & 0.009 & 0.000 & 0.019 & 0.001 &0.012 & 0.001 \\
			\hline
			\multirow{2}{*}{Uncer} & None & 0.114 & 0.005 & 0.201 & 0.015 & 
			0.418 & 0.017 \\
			\cline{2-8}
			& Purifier & 0.006 & 0.000 & 0.007 & 0.001 & 0.006 & 0.001 \\
			\hline
		\end{tabularx}
	}
 \vspace{-0.5cm}
\end{table}

\codename reshapes the input confidence score vectors according to the pattern 
of the learned non-member samples.
We examine the indistinguishability of the confidence scores on members and 
non-members by plotting the histogram of the target classifier's confidence in 
predicting the correct class and the prediction uncertainty in Figure~\ref{fig:prediction_acc}.
The prediction uncertainty is measured as the normalized entropy 
$\frac{-1}{\log (k)} \sum_i \hat{\vec{y}_i}\log (\hat{\vec{y}_i})$ of the 
confidence vector $\vec{y} = F(\vec{x})$, where $k$ is the number of classes.
As Figure~\ref{fig:prediction_acc} 
shows, 
\codename can reduce the gap between the two curves.
Similar curves can be obtained on CIFAR10 and FaceScrub530 classifiers. We attach them to Appendix. 
We also report the maximum gap and the average gap between the curves 
in Table~\ref{tb:measurement}.
The results show that our approach can significantly reduce both the maximum 
and average gaps between the target classifier's confidence in predicting the 
correct class as well as the prediction uncertainty on its members versus 
non-members.
This demonstrates that \codename successfully reduces the individual 
differences between members and non-members.

\noindent \underline{\textbf{Statistical Indistinguishability of Purified 
		Confidence}}

\begin{figure}[t]
        \vspace{-0.2cm}
        \setlength{\belowcaptionskip}{-0.65cm}
        \setlength{\abovecaptionskip}{0.15cm} 
        \subfigbottomskip=1pt
	\centering   
	\subfigure[\scriptsize Original member latents.]{
		\includegraphics[width=0.4\linewidth]{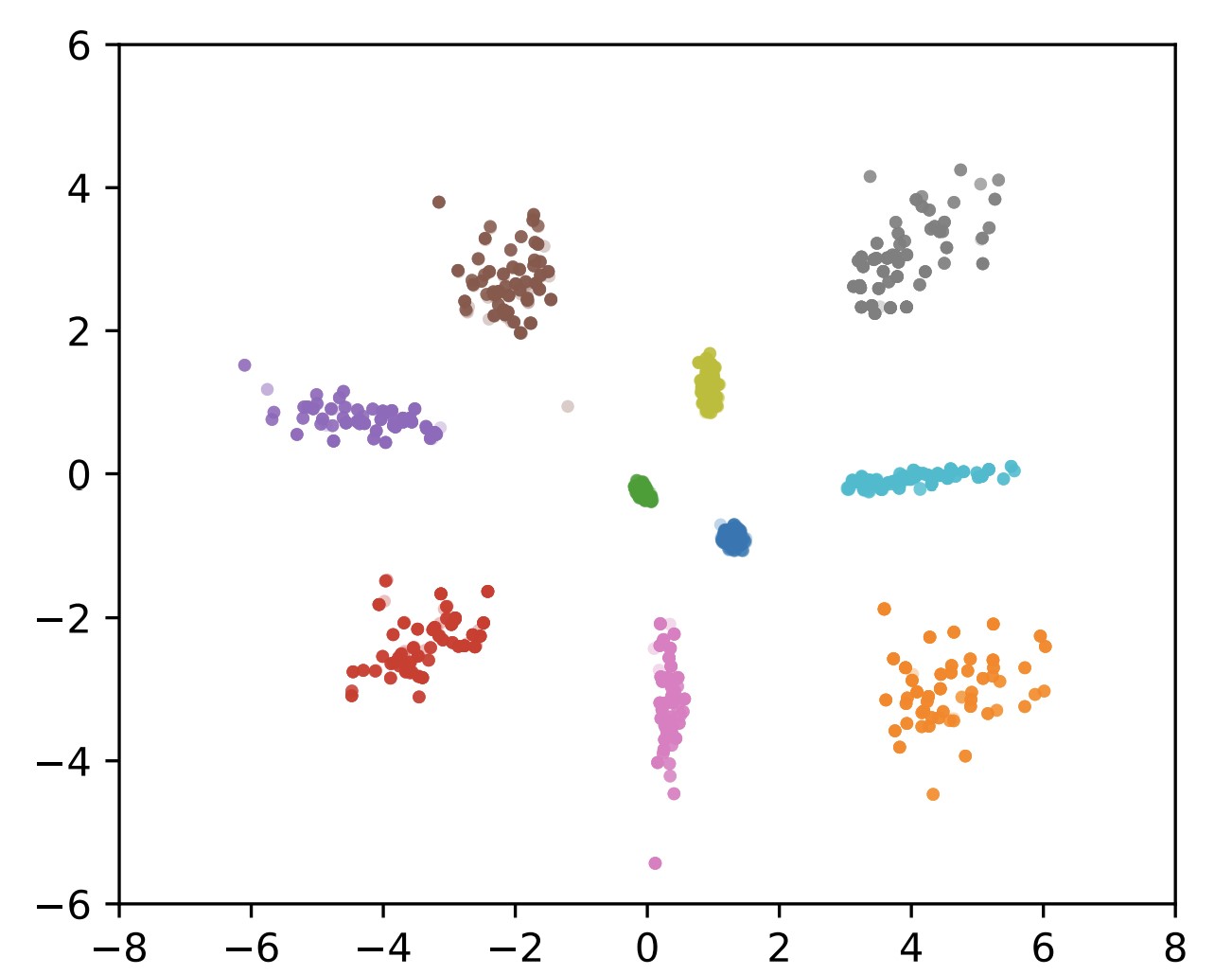}
	}
	\subfigure[\scriptsize Original non-member latents.]{
		\includegraphics[width=0.46\linewidth]{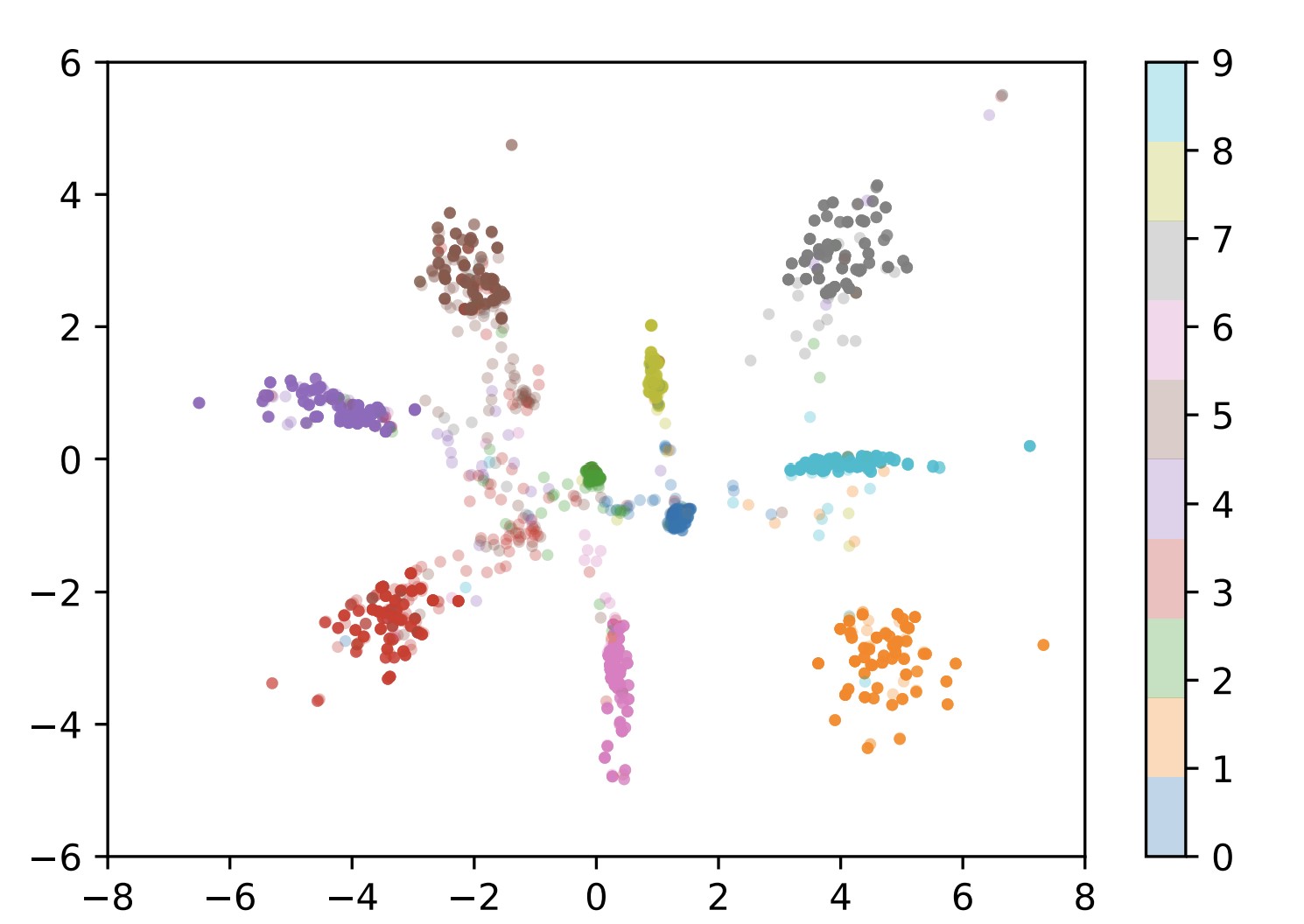}
	}
	\subfigure[\scriptsize Purified member latents.]{
		\includegraphics[width=0.4\linewidth]{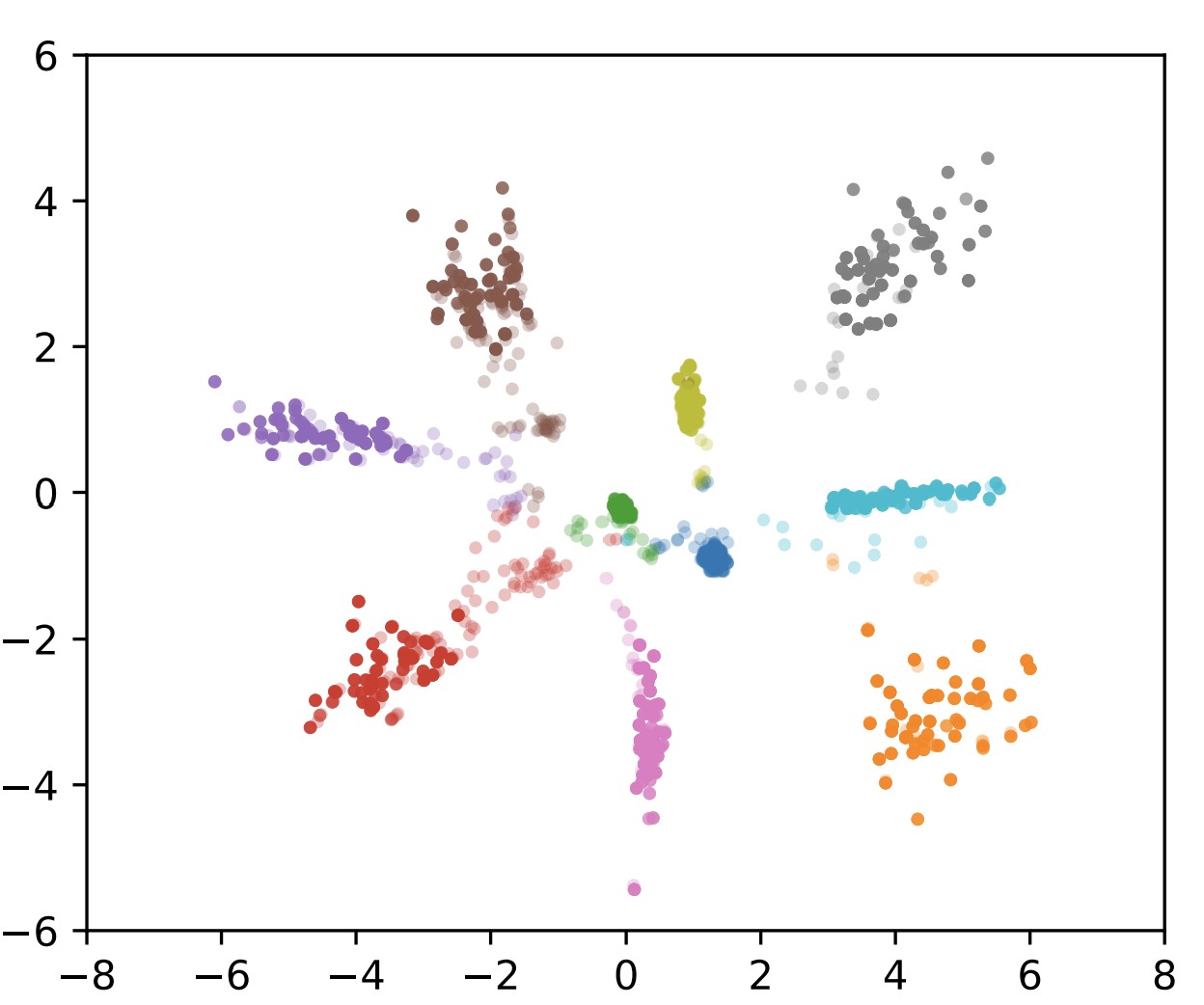}
	}
	\subfigure[\scriptsize Purified non-member latents.]{
		\includegraphics[width=0.46\linewidth]{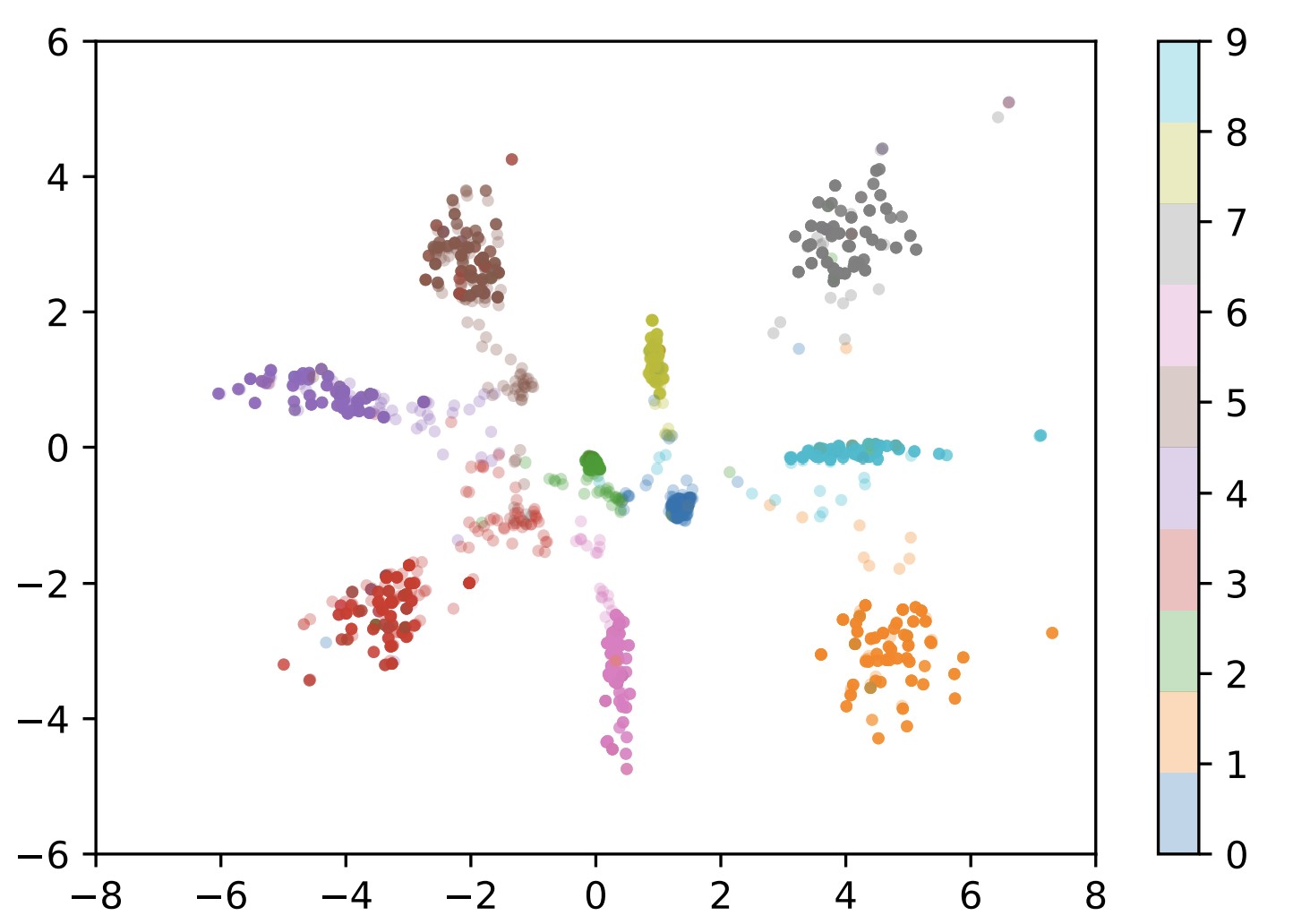}
	}
	\caption{The statistical distribution of latent vectors on the CIFAR10 dataset. Different colors stand for latent vectors with 
		different labels. (a) and (b) depict latent vectors of the original member 
		and non-member confidence score vectors;
		(c) and (d) depicts latent vectors of member and non-member confidence 
		score vectors with \codename defended.}
	\label{fig:latent_space_purifier}
\end{figure}

We present the statistical distribution of confidence score vectors in the encoder latent space of the \textit{confidence reformer}. Figure~\ref{fig:latent_space_purifier} visibly displays the differences  on the CIFAR10 dataset between 
members and non-members in 
the latent space.
As illustrated in the first row, latent vectors of the 
members tend to cluster together according to their labels, while those of 
non-members are more scattered in the map.  
The second row of Figure~\ref{fig:latent_space_purifier} also shows the statistical distribution of members and 
non-members processed with \codename in the 
latent space. When processed with \codename, Gaussian noises are added to the confidence score vectors, making the clustered member latent vectors to be more
scattered on the latent space. 
This demonstrates that \codename can reduce the statistical differences
between members and non-members while preserving semantic utility.

\noindent \underline{\textbf{Label Indistinguishability of Purified 
		Confidence}}

\codename uses \textit{label swapper} to identify and swap the predicted labels of members. 
\textit{label swapper} incurs 
negligible reduction of test accuracy. At the same time, swapping the labels of the member samples reduces the training accuracy so that the gap between the accuracy of member and non-members is minimized.
This is shown in 
Table~\ref{tb:bigtable}, where the 
training accuracy of the model is close to the test accuracy. Many label-only 
membership inference attacks are less effective under \codename with \textit{label swapper}.
This reflects that purified member confidence vectors 
are less distinguishable from those of the non-members in terms of label.

\section{Discussion}

Assuming the reference data are considered as members, we present the inversion error and the inference accuracy
on the reference set for each defense and attach the results to the Appendix.  
The Results show that 
\codename can still preserve the defense effect against the adversarial model inversion attack and the membership inference attack.

We also investigate the effect of the \codename’s training data by using different in-distribution and out-of-distribution data
to train \codename. The results show that \codename can still mitigate the attacks, but at the cost of sacrificing the utility significantly when using out-of-distribution data. We attach the results to the Appendix.

Furthermore, we investigate the effectiveness of \codename to detect noisy members and attach the result to Appendix. It shows that \codename can accurately detect the members with noise $\left \| \eta \right \|_{\infty}<1e-10$ on FaceScrub530 dataset.

\section{Related Work}

\textbf{Data Inference Attacks.}
In data inference attacks, the attacker aims at inferring information about the data that the target model operates on.
Xiao et al.~\cite{xiao_adversarial_2019} studied the adversarial reconstruction problem.
They studied the prediction model which outputs 40 binary attributes. 
Our paper, on the contrary, studies black-box classifiers whose output is constrained by a probability distribution. 
Jia and Gong~\cite{jia_attriguard_2018} proposed the adversarial formulation for privacy protection. They aimed at protecting the privacy of users' sensitive attributes from being inferred from their public data. Our work investigates inference attacks that leverage prediction results of machine learning models to infer useful information about the input data.

\textbf{Secure \& Privacy-Preserving Machine Learning.}
A number of studies made use of trusted hardware and cryptographic computing to provide secure and privacy-preserving training and use of machine learning models. 
These techniques include homomorphic encryption, garbled circuits and secure multi-party computation on private data~\cite{liu_oblivious_2017, bonawitz_practical_2017, phong_privacy-preserving_2018, dowlin_cryptonets_2016, mohassel_secureml_2017, dwork_privacy-preserving_2018} and secure computing using trusted hardware~\cite{ohrimenko_oblivious_2016, juvekar_gazelle_2018}.
Although these methods protect sensitive data from direct observation by the attacker, they do not prevent information leakage via the model computation. 
\section{Conclusion}

In this paper, we propose \codename to defend data inference attacks.
\codename learns the pattern of non-member confidence score vectors and 
purifies confidence score vectors to this pattern. It makes member 
confidence score vectors indistinguishable from non-members in terms of 
individual shape, statistical distribution and prediction label. 
Our experiments show that \codename is effective and efficient in mitigating existing data inference attacks, outperforming previous defense methods,
while imposing negligible utility loss. 

\clearpage
\section{Ethics Statement}
The code for \codename is available at \url{https://github.com/wljLlla/Purifier_Code}.
\section{Acknowledgments}
This work was supported in part by National Key R{\&}D Program of China (2020AAA0107700),  by National Natural Science Foundation of China (62102353, 62227805), by National Key Laboratory of Science and Technology on Information System Security (6142111210301), by State Key Laboratory of Mathematical Engineering and Advanced Computing, and by Key Laboratory of Cyberspace Situation Awareness of Henan Province (HNTS2022001). We would like to thank Dingkun Wei, Jingjing Wang, Zijing Hu and Yanqing Liu for their implementation of some experiments.

\appendix
\clearpage
\section{Appendix}

\subsection{Experimental Setup}

\noindent \underline{\textbf{Datasets}}

We use CIFAR10, Purchase100,  
FaceScrub530 and UTKFace datasets
which are widely adopted in previous studies on 
membership 
inference attacks, attribute inference attacks and model inversion attacks.
Additionally, we also evaluate the effectiveness of \codename against memership inference attacks on CIFAR100, Texas and Location. 

\noindent \textbf{CIFAR10.}
It is a machine learning benchmark dataset for evaluating image recognition 
algorithms with 10 classes. 
It consists of 60,000 color images, each of size 32 x 32.
The dataset has 10 
classes, where each class represents an object (e.g., airplane, car, etc.).

\noindent \textbf{Purchase100.}
This dataset is based on Kaggle's ``acquired valued shopper'' challenge.%
\footnote{https://www.kaggle.com/c/acquire-valued-shoppers-challenge/data}
We used the preprocessed and simplified version of this 
dataset.
It is composed of 197,324 data records and each data record has 600 binary 
features. The dataset is clustered into 100 classes.

\noindent \textbf{FaceScrub530.}
This dataset consists of URLs for 100,000 images of 530 individuals.
We obtained the preprocessed and simplified version of this dataset from 
 which has 48,579 facial images and each image is 
resized to 64 $\times$ 64.

\noindent \textbf{UTKFace.}
This dataset consists of URLs for 22000 images of individuals. We train the classifier to classify the genders and use the race as the sensitive attribute in our experiments.

\noindent \textbf{CIFAR100.}
It is a machine learning benchmark dataset for evaluating image recognition 
algorithms with 10 classes. 
It consists of 60,000 color images, each of size 32 x 32.
The dataset has 100
classes and each classes has 600 images.

\noindent \textbf{Texas.}
We use the same data as previous studies, which cluster the data with 6169 attributes to 100 classes.

\noindent \textbf{Location.}
This dataset is based on the publicly
available set of mobile users’ location “check-ins” in the
Foursquare social network, restricted to the Bangkok area
and collected from April 2012 to September 2013. The record of Location has 446 attributes and   has clustered into 30 classes.

\noindent \underline{\textbf{Target Classifier}}

We use the same model architectures as in previous 
work.
For CIFAR10 and CIFAR100 datasets, we use DenseNet121. 
We train our classifier with SGD optimizer for 
350 epochs with the learning rate 0.1 from epoch 0 to 150, 0.01 from 150 to 250, 
and 0.001 from 250 to 350. The classifier is regularized with 
$L_2$ regularization (weight decay parameter 5e-4).
For Purchase100 dataset, we use the same model and training strategy as in previous work
to train the target classifier. It is a 4-layer fully 
connected neural network.
For FaceScrub530 dataset, we use the same conventional neural network and 
the same training strategy as in previous work to train the target 
classifier.
For UTKFace dataset, we use the same neural network and 
the same training strategy as used in FaceScrub530 dataset, except that the output layer dimension is changed to 2.  
For Texas dataset, we use a 4-layer fully 
connected neural network with the Tanh as the activation function.
For Location dataset, we use a 3-layer fully 
connected neural network with the Tanh as the activation function.

\noindent \underline{\textbf{\codename}}

We use CVAE to implement the \textit{confidence reformer} $G$. It has the layer size of [20, 32, 
64, 128, 2, 128, 64, 32, 20] for CIFAR10, [200, 128, 256, 512, 20, 512, 256, 
128, 100] for Purchase100, Texas and location, [1060, 512, 1024, 2048, 100, 2048, 1024, 512, 
1060] for FaceScrub530 and CIFAR100. The layer size of [4, 32, 
64, 128, 2, 128, 64, 32, 2] for UTKFace, . 
We use ReLU and batch normalization in hidden 
layers.
We train \codename on Purchase100 dataset for 150 epochs, CIFAR10, Texas and Location  datasets for 
100 epochs, FaceScrub530, UTKFace and CIFAR100 datasets for 300 epochs. 
We use Adam optimizer with the learning rate 0.01 for CIFAR10, 0.0001 for 
Purchase100, Texas and Location, 0.0005 for Facescrub530 and CIFAR10 and 0.0001 for UTKFace.

\noindent \underline{\textbf{Existing Attacks}}

In our experiments, we implement the following data inference attacks.

\noindent \textbf{NSH attack}.
This attack assumes that the attacker knows both the membership labels and ground truth of $D_{aux}$, and thus no shadow is trained.
The membership classifier makes use of both the confidence score vector and the ground truth of a data sample to predict its membership.

\noindent \textbf{Mlleaks attack}.
In this attack, the attacker knows the ground truth of $D_{aux}$ but does not know their membership labels. Therefore, a shadow model is required to replicate the target model. 
We use half of $D_{aux}$ to train the shadow model which has the same architecture as the target classifier, and use the whole $D_{aux}$ to train the membership classifier with their labeled membership information in terms of the shadow model.
The membership classifier is a multi-layer perception with a 128-unit hidden layer and a sigmoid output layer.
All weights were initialized with normal distribution with a mean of 0 and standard deviation of 0.01, and all biases are initialized to 0. 
We use the Adam optimizer with learning rate 0.001. The number of training epochs is set to 50 for each dataset.

\noindent \textbf{Adaptive attack}. 
This is an adaptive version of the Mlleaks attack, where the attacker is 
assumed to know all the details of the defender's \codename and its training 
data $D_2$. Hence, the attacker trains the same \codename and appends it to the 
shadow model. The membership classifier is then trained on the purified 
confidence score vectors.

\noindent \textbf{BlindMI attack}. We consider BlindMI-DIFF-w/, where the attacker is assumed to know 
the soft label and the ground truth of the target dataset. We use sobel to generate non-member samples. The size of the non member dataset $|S_{nonmem}|$ is 20. 

\noindent \textbf{Label-only attack}. In Gap Attack, the attacker is assumed to have the ground truth of the data sample and 
predicts that it is a member if and only if the target classifier gives the 
correct label on the sample. 
In the Transfer attack, auxiliary dataset $D_{aux}$ is re-labeled by querying the target model. The shadow models that we adopt share the same architecture with the target model. Three different thresholds are chosen in previous work, and we consider the one that gives the best result on $D_{aux}$. In the Boundary attack, we adopt HopSkipJump noise, with a total evaluation of {15000 per sample}, which ensures the attack performance is stable. Similarly, we report the results on the $L_2$ norm. Due to high query-complexity, the results are reported on a subset that consists of 1000 samples. 

\noindent \textbf{Adversarial Model inversion attack}.
The attacker trains an inversion model on $D_{aux}$ to perform the model inversion attack.

\noindent \textbf{Attribute inference attack}.
The attacker trains a classification on $D_{aux}$ to predict the additional sensitive attribute taking the confidence vectors as input. The classifier is a multi-layer perceptron.
All weights were initialized with normal distribution with a mean of 0 and standard deviation of 0.01, and all biases are initialized to 0. 
We use the Adam optimizer with learning rate 0.00001. The number of training epochs is set to 1500 for the UTKFace dataset.	

\begin{table}[h]
	\centering
        \tiny
	\caption{Data allocation. 
 A dataset is divided into training set $D_{1}$ of the target classifier, reference set $D_{2}$ and test set $D_{3}$. 
		In membership inference attack, we assume that the attacker has access to a subset $D^{A}$ of $D_{1}$ and a subset $D^{'A}$ of $D_{3}$.
  }
	\label{tb:datasplit}
	\resizebox{\columnwidth}{!}{
		\begin{tabular}{c|l|l|l|c|r}
			\hline
			Dataset &  \multicolumn{1}{c|}{$D_{1}$} & \multicolumn{1}{c|}{$D_{2}$} & \multicolumn{1}{c|}{$D_{3}$} & \multicolumn{1}{c|}{$D^{A}$} & \multicolumn{1}{c}{$D^{'A}$} \\
			\hline
			CIFAR10 & 50,000 & 5,000 & 5,000 & 25,000 & 2,500\\
			Purchase100 & 20,000 & 20,000 & 20,000 & 10,000 & 10,000\\
			FaceScrub530 & 30,000 & 10,000 & 8,000 & 15,000 & 4,000\\
   			UTKFace & 12,000 & 5,000 & 5,000 & 6,000 & 2,500\\
                CIFAR100 & 50,000 & 5,000 & 5,000 & 25,000 & 2,500\\
                Texas & 10,000 & 10,000 & 10,000 & 5,000 & 5,000\\
                Location & 1,600 & 1,600 & 1,600 & 800 & 800\\
			\hline
		\end{tabular}
	}
\end{table}

Table~\ref{tb:datasplit} presents the data allocation in our experiments.
We divide each dataset into the target classifier's training set $D_1$, the reference set $D_2$ and the test set $D_3$. 
In membership inference and attribute inference attack,
we assume that the attacker has access to a subset $D^{A}$ of $D_1$ and a subset $D^{'A}$ of $D_3$ to form its auxiliary dataset $D_{aux}$.
We use the remaining data in $D_1$ and $D_3$ to test the membership inference and attribute inference accuracy.
We use $D_2$ as the reference dataset for defenses that require such a dataset.
In the model inversion attack, for the FaceScrub530 classifier, the attacker uses a CelebA\footnote{ICCV 2015 Deep Learning Face Attributes in the Wild}
dataset to train the inversion model. For other classifiers, the attacker samples 80\% from $D_1$, $D_2$ and $D_3$ respectively to train the inversion model, and uses the remaining 20\% data to test the inversion error.

\noindent \underline{\textbf{Metrics}}

We use the following metrics to measure the utility, defense performance and efficiency of a defense method.

\noindent \textbf{Classification Accuracy.} It is measured on the training set and the test set of the target classifier. It reflects how good the 
target classifier is on the classification task. 

\noindent \textbf{Utility Error.} It is measured on the training set 
$D_1$ and the test set $D_3$ of the target classifier. It reflects how good the 
target classifier is on the classification task.

\noindent \textbf{Inference Accuracy.} This is the classification accuracy of 
the attacker's attack model in predicting the membership and additional sensitive attribute of input samples. 

\noindent \textbf{Inversion Error.} We measure the inversion error by computing the mean squared error between the original input sample and the reconstruction. 
For the FaceScrub530 classifier, it is measured on $D_1$ and $D_3$. For other classifiers, it is measured on the 20\% of $D_1$ and $D_3$ respectively.

\noindent \textbf{Efficiency.} We measure the efficiency of a defense method by 
reporting its training time and testing time relative to the original time required by the target 
classifier.

\subsection{Experiment of Adversarial Model Inversion in MNIST Dataset}
\begin{figure}[t]
	\centering   
	\subfigure{
		\includegraphics[width=1\linewidth]{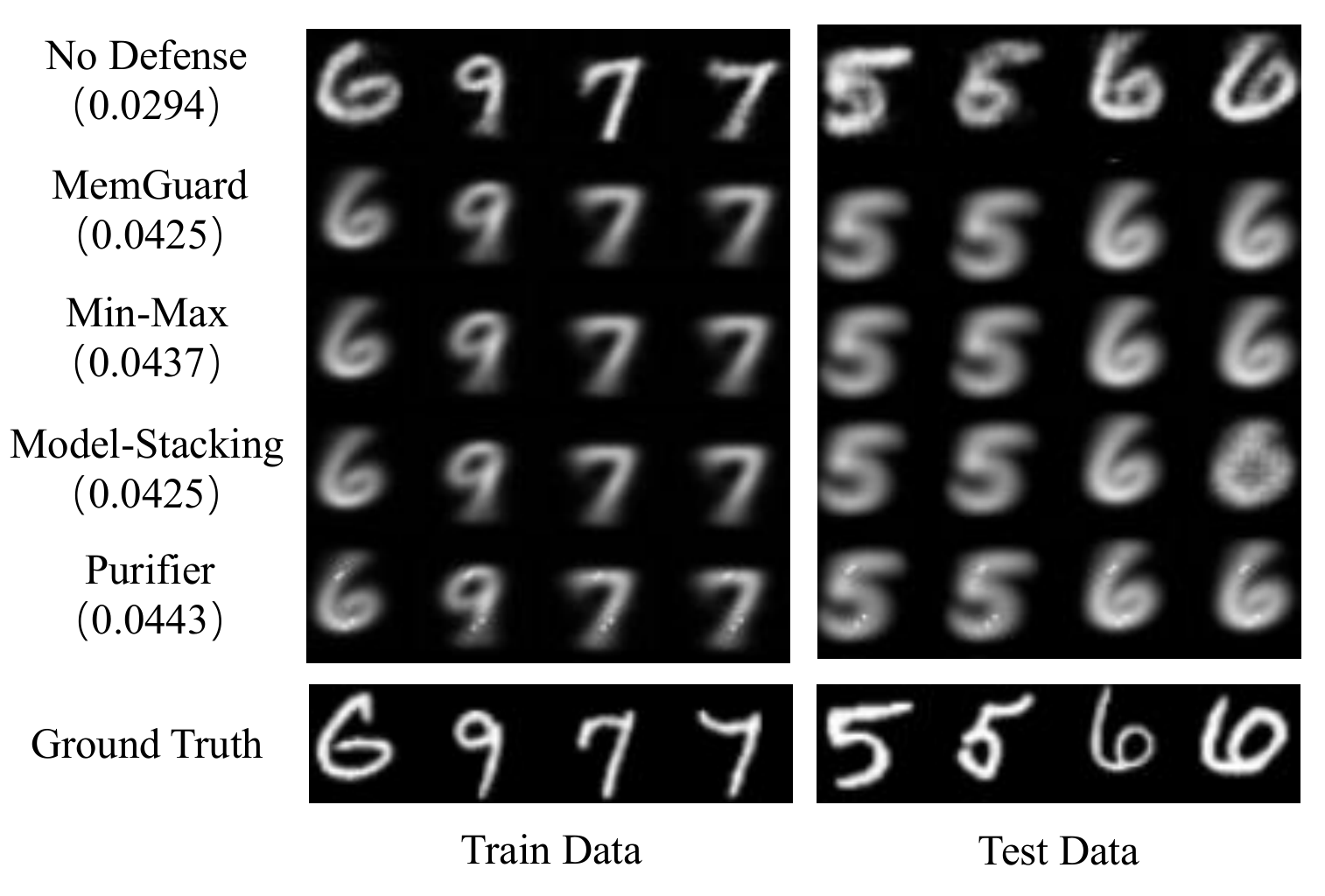}
	}
	\caption{Model inversion attack against the MNIST classifier 
		defended by different approaches. Numbers on the left indicate inversion errors}
	\label{fig:mnist}
\end{figure}

We investigate the \codename against the adversarial model inversion attack on the MNIST dataset and plot the reconstructed images in Figure~\ref{fig:mnist}. We report the inversion error at
the left of Figure~\ref{fig:mnist} to qualify the inversion performance. As
illustrated in Figure~\ref{fig:mnist}, the attacker is able to reconstruct nearly
identical images when no defense is used. However, it is much
more difficult for the attacker to reconstruct the image from the
purifier-defended model. The reconstructed images lose a lot
of details compared with the original one, only representing a
blurred image of their classes.   

\subsection{Distribution of the target classifier’s prediction
uncertainty and confidence in correct class}

\begin{figure}[t]
	\centering
	\begin{minipage}[b]{1\linewidth}
		\centering
		\subfigure{
			\includegraphics[width=0.5\linewidth]{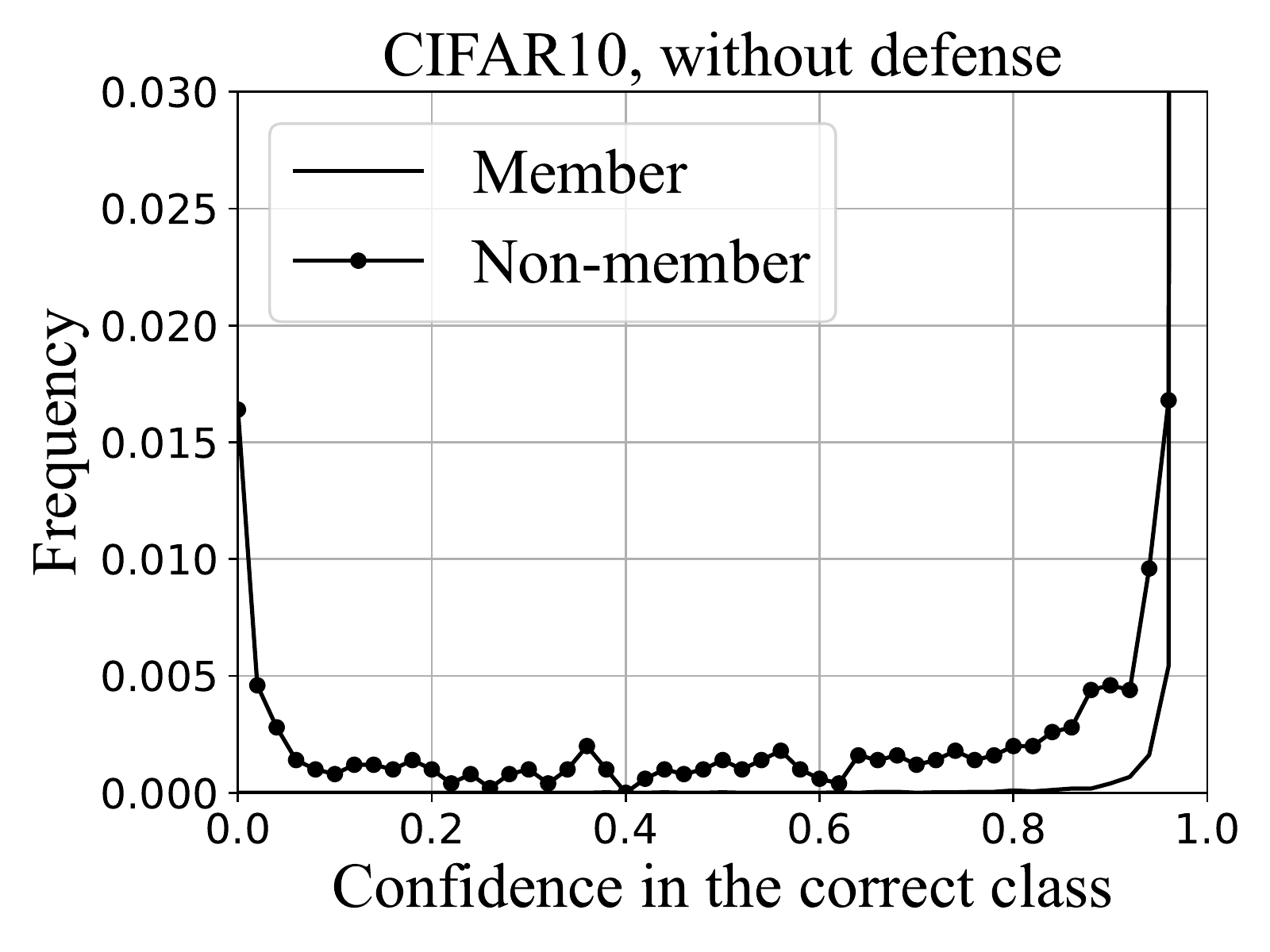}
			\includegraphics[width=0.5\linewidth]{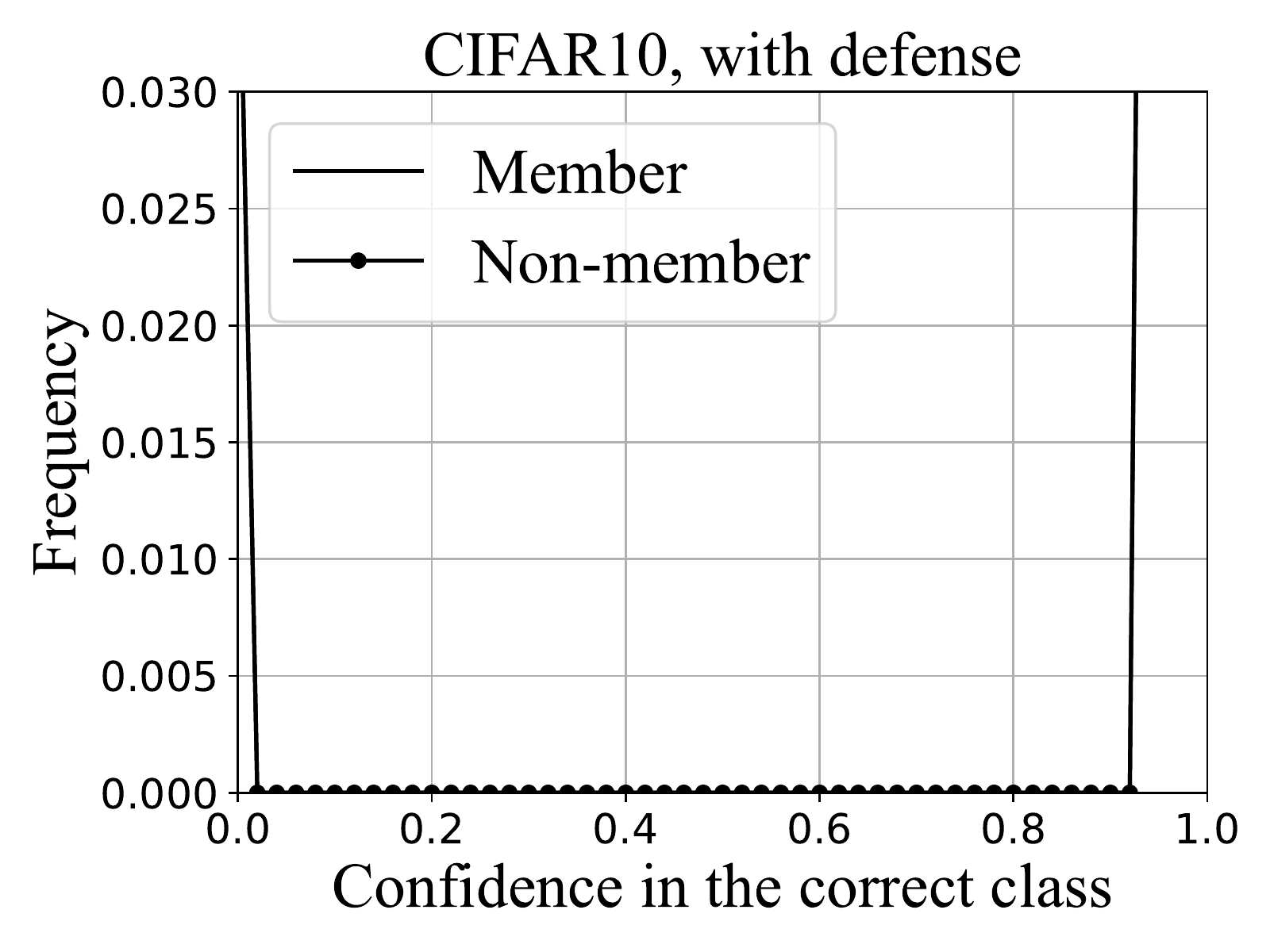}
		}
		
		\subfigure{
			\includegraphics[width=0.5\linewidth]{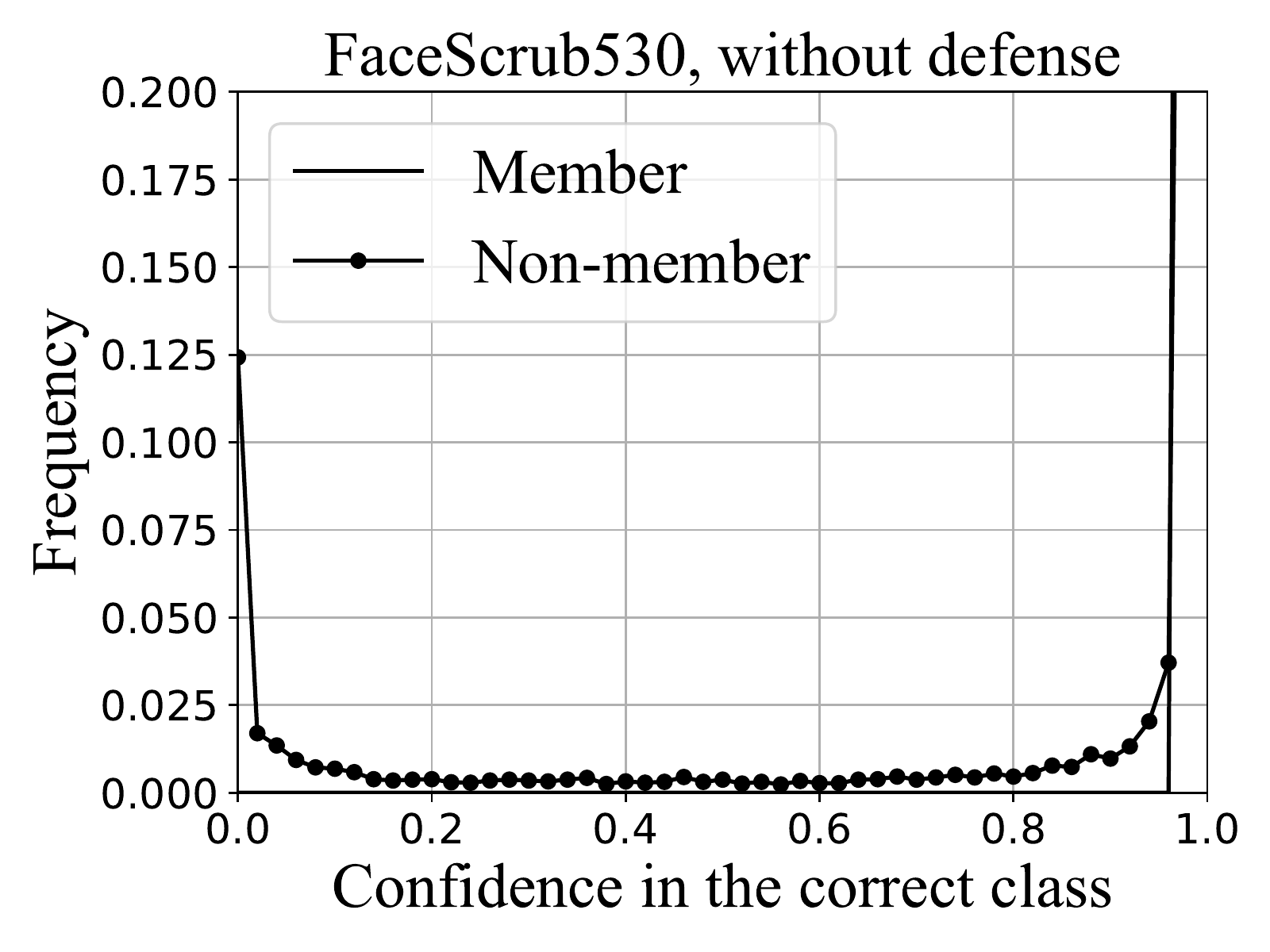}
			\includegraphics[width=0.5\linewidth]{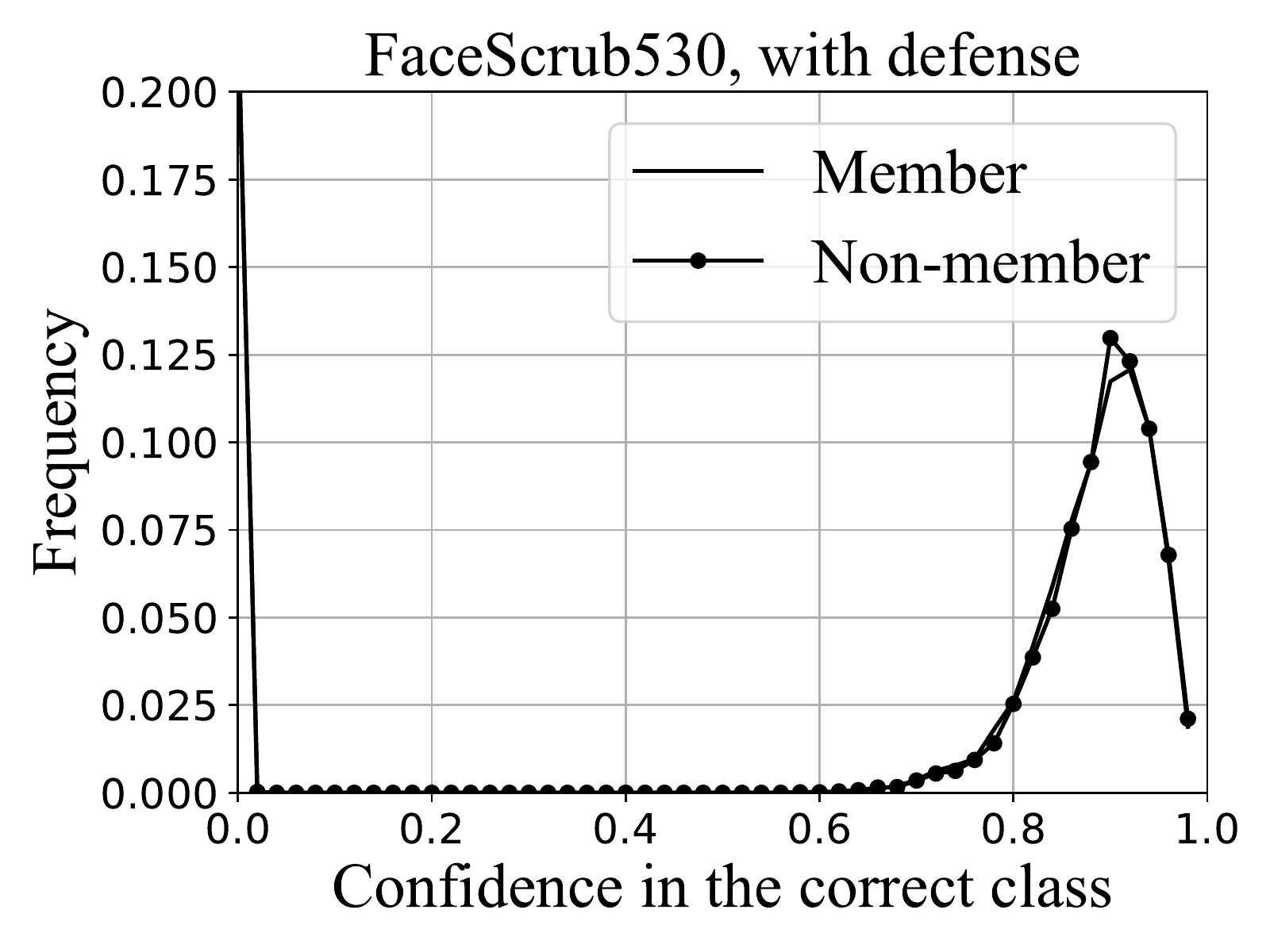}
		}
	\end{minipage}
	
	\caption{Distribution of the target classifier's confidence in predicting the correct class on members and non-members of its training set.}
	\label{fig:prediction_acc_other}
\end{figure}

\begin{figure}[h!]
	\centering
	\begin{minipage}[b]{1\linewidth}
		\centering
		\subfigure{
			\includegraphics[width=0.5\linewidth]{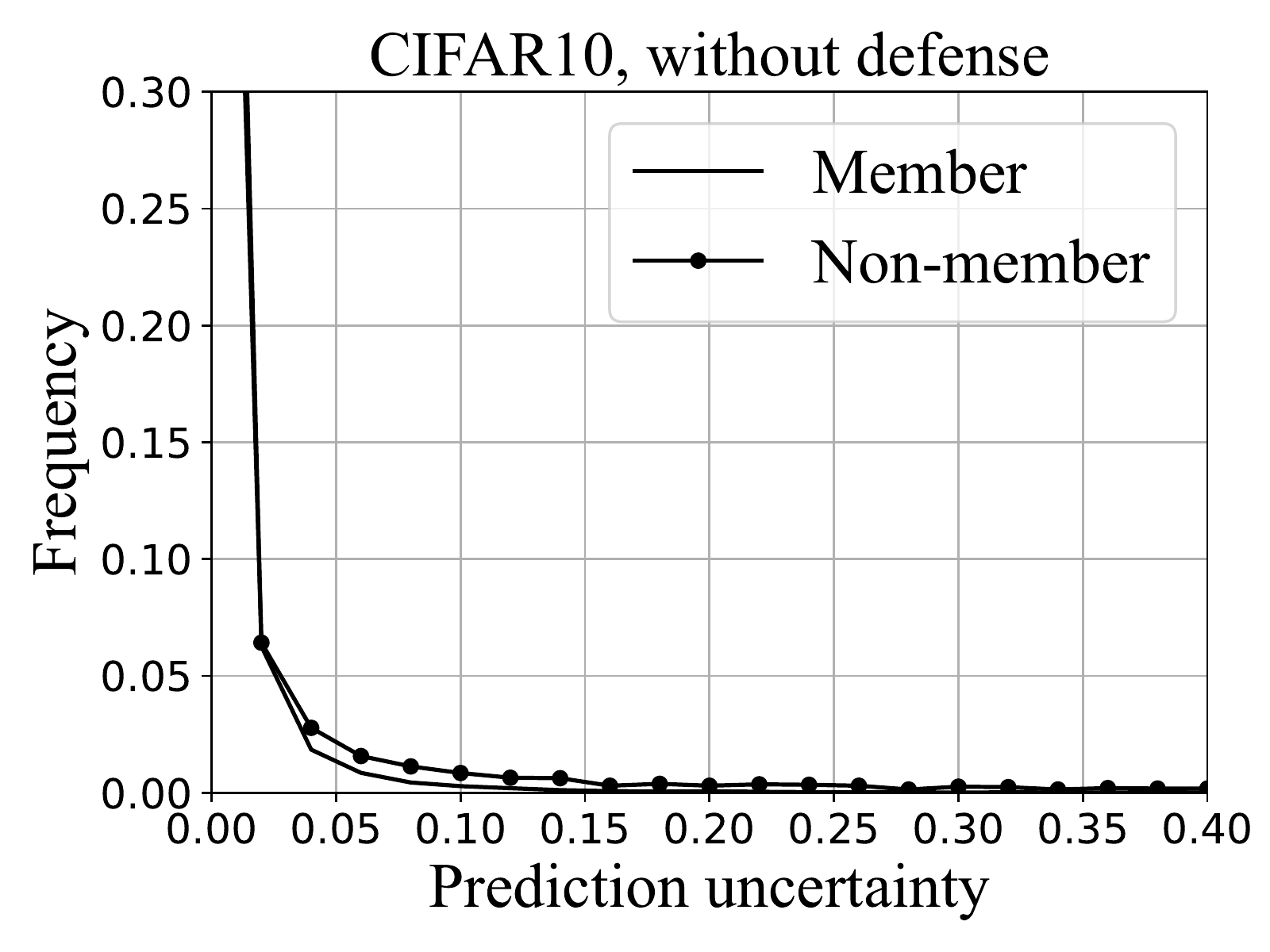}
			\includegraphics[width=0.5\linewidth]{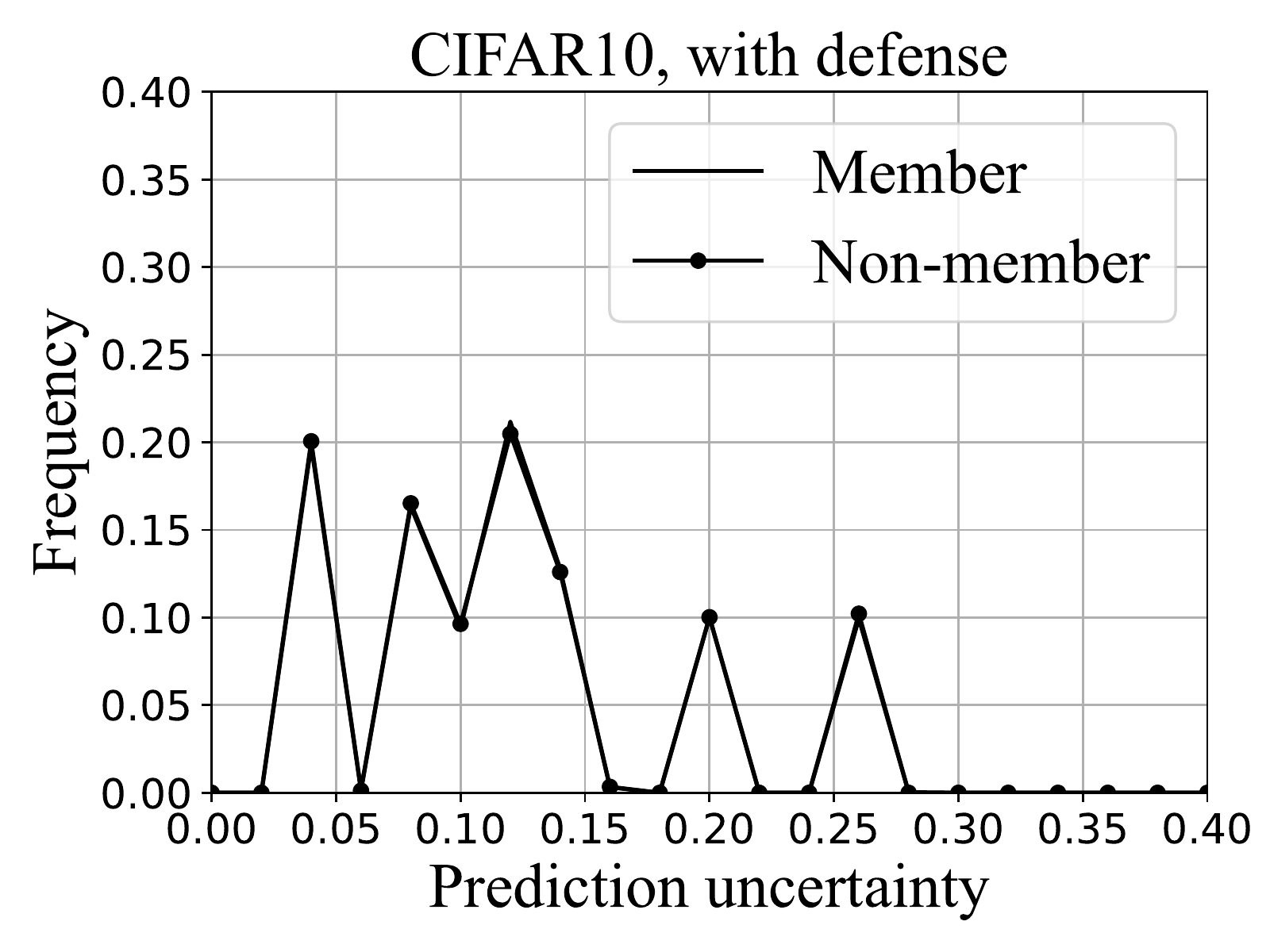}
		}
		
		\subfigure{
			\includegraphics[width=0.5\linewidth]{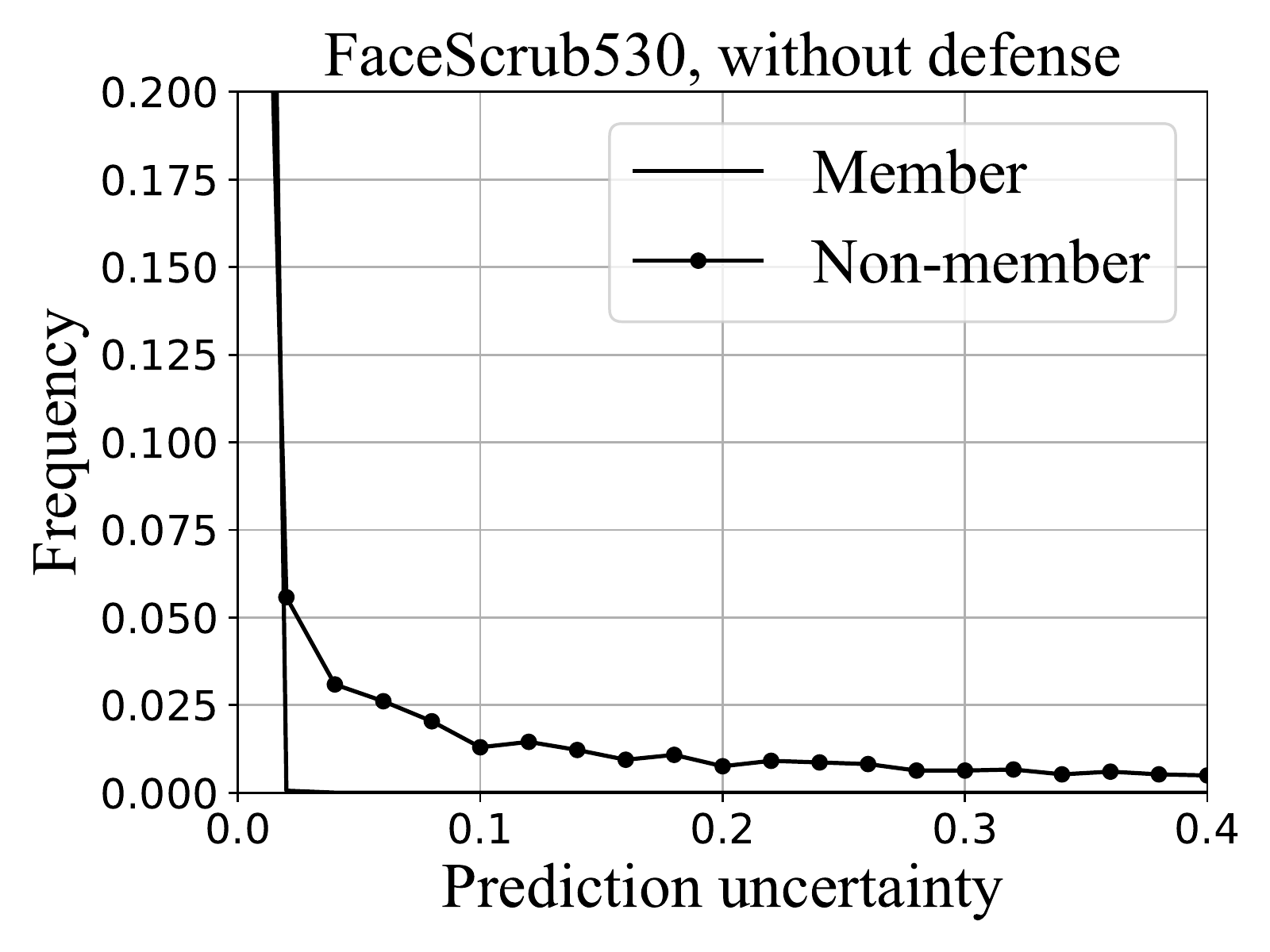}
			\includegraphics[width=0.5\linewidth]{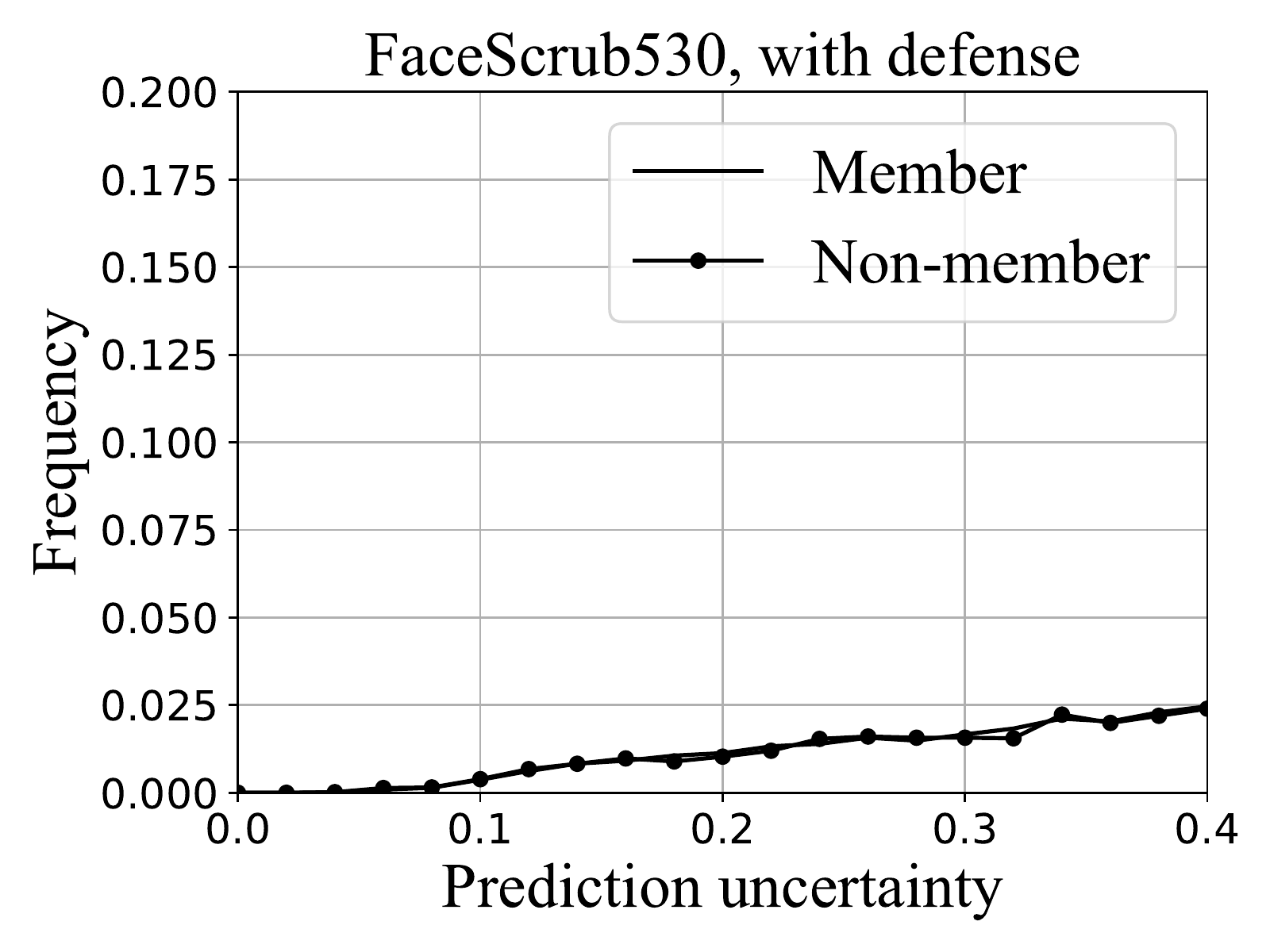}
		}
	\end{minipage}
	\caption{Distribution of the target classifier's prediction uncertainty on members and non-members of its training set. The uncertainty is measured as the normalized entropy of the confidence score vector.}
	\label{fig:uncertainty_other}
\end{figure}

We examine the indistinguishability of the confidence scores on members and 
non-members by plotting the histogram of the target classifier's confidence in 
predicting the correct class and the prediction uncertainty in CIFAR10 and FaceScrub530 datasets. 
We show the results in Figure~\ref{fig:prediction_acc_other} and Figure~\ref{fig:uncertainty_other}.
As Figure~\ref{fig:prediction_acc_other} and Figure~\ref{fig:uncertainty_other} 
show, 
\codename can also reduce the gap between the two curves in CIFAR10 and FaceScrub530 datasets.
This demonstrates that PURIFIER also successfully reduces the
individual differences between members and non-members in CIFAR10 and FaceScrub530 datasets.

\begin{table*}[t]
        \scriptsize
  	\setlength{\tabcolsep}{0.4\tabcolsep}
	\centering
	\caption{Defense performance of \codename against SELENA on additional datasets. Results of Transfer attack and Boundary attack are reported in AUC. Note 
	that the N.A. means that setting is not applicable. the code of Boundary attack can not be used on non-image dataset.
 }
 \label{tb:new_dataset}
 \resizebox{\linewidth}{!}{
\begin{tabular}{c|c|c|c|c|c|c|c|cc}
\toprule
\multirow{2}{*}{Dataset}                        & \multirow{2}{*}{Defense} & \multirow{2}{*}{Training acc.} & \multicolumn{2}{l|}{\multirow{2}{*}{Test acc.}} & \multirow{2}{*}{NSH Attack} & \multirow{2}{*}{Mlleaks Attack} & \multirow{2}{*}{Adaptive Attack} & \multicolumn{2}{l}{Label only Attacks}   \\ \cline{9-10} 
                                                &                          &                                & \multicolumn{2}{l|}{}                           &                             &                                 &                                  & \multicolumn{1}{l|}{transfer} & boundary \\ \hline
\multicolumn{1}{c|}{\multirow{3}{*}{CIFAR100}} & None               & 100\%                          & \multicolumn{2}{l|}{69.98\%}                    & 76.98\%                     & 73.78\%                         & 73.78\%                          & \multicolumn{1}{l|}{0.5980}   & 0.6668   \\ \cline{2-10} 
\multicolumn{1}{c|}{}                          & SELENA                   & 78.00\%                        & \multicolumn{2}{l|}{62.10\%}                    & 50.32\%                     & 50.42\%                         & 51.38\%                          & \multicolumn{1}{l|}{0.5188}   & 0.5008   \\ \cline{2-10} 
\multicolumn{1}{c|}{}                          & Purifer                  & 70.02\%                        & \multicolumn{2}{l|}{69.98\%}                    & 50.01\%                     & 50.15\%                         & 51.02\%                          & \multicolumn{1}{l|}{0.5120}   & 0.4975   \\ \hline
\multirow{3}{*}{Texas}                          & None               & 79.17\%                        & \multicolumn{2}{l|}{50.91\%}                    & 66.37\%                     & 58.93\%                         & 58.93\%                          & \multicolumn{1}{l|}{0.5431}   & N.A.     \\ \cline{2-10} 
                                                & SELENA                   & 56.70\%                        & \multicolumn{2}{l|}{47.21\%}                    & 54.10\%                     & 51.04\%                         & 51.51\%                          & \multicolumn{1}{l|}{0.5086}   & N.A.     \\ \cline{2-10} 
                                                & Purifer                  & 51.01\%                        & \multicolumn{2}{l|}{50.91\%}                    & 51.29\%                     & 50.00\%                         & 51.18\%                          & \multicolumn{1}{l|}{0.5028}   & N.A.     \\ \hline
\multirow{3}{*}{Location}                       & None               & 100\%                          & \multicolumn{2}{l|}{60.44\%}                    & 82.37\%                     & 84.00\%                         & 84.00\%                          & \multicolumn{1}{l|}{0.5893}   & N.A.     \\ \cline{2-10} 
                                                & SELENA                   & 77.90\%                        & \multicolumn{2}{l|}{55.25\%}                    & 54.40\%                     & 51.00\%                         & 53.25\%                          & \multicolumn{1}{l|}{0.5012}   & N.A.     \\ \cline{2-10} 
                                                & Purifer                  & 60.45\%                        & \multicolumn{2}{l|}{60.43\%}                    & 51.75\%                     & 50.31\%                         & 51.41\%                          & \multicolumn{1}{l|}{0.5015}   & N.A.     \\ \bottomrule
\end{tabular}
    }
\end{table*}

\subsection{Experiments on Other Three Datasets}
We evaluate the effectiveness of \codename against various membership inference attacks on CIFAR100, Texas and Location datasets. Table~\ref{tb:new_dataset} shows that \codename is able to reduce the performance of the membership inference attacks significantly on the additional three datasets and outperforms SELENA in most of cases. For instance, \codename can reduce the accuracy of NSH attack on Texas dataset from 66.37\% to 51.29\% while SELENA can only reduce it from 66.37\% to 54.10\%.   

\begin{table}[t]
	\caption{Results of \codename against attackers with the more powerful shadow models.}
	\label{tb:powerful_shadow}
\begin{tabular}{c|c|c|c}
\toprule
\multirow{2}{*}{Dataset}                       & \multirow{2}{*}{Defense} & \multirow{2}{*}{NSH Attack} & \multirow{2}{*}{Mlleaks Attack} \\
                                               &                          &                             &                                 \\ \hline
\multicolumn{1}{c|}{\multirow{2}{*}{CIFAR10}} & None                     & 58.46\%                     & 70.20\%                         \\ \cline{2-4} 
\multicolumn{1}{c|}{}                         & Purifer                  & 50.67\%                     & 50.01\%                         \\ \hline
\multirow{2}{*}{Purchase100}                   & None                     & 88.27\%                     & 68.50\%                         \\ \cline{2-4} 
                                               & Purifer                  & 52.74\%                     & 51.00\%                         \\ \hline
\multirow{2}{*}{FaceScrub530}                  & None                     & 70.30\%                     & 73.46\%                         \\ \cline{2-4} 
                                               & Purifer                  & 50.22\%                     & 50.66\%                         \\ \bottomrule
\end{tabular}
\end{table}

\subsection{Effectiveness of \codename against Attackers with More Powerful Shadow Models}
In order to study the effectiveness of \codename against stronger attackers, we evaluate the performance of \codename against NSH attack and Mlleaks attack which use the same number of the data as the target model to train a more powerful shadow model. As Table~\ref{tb:powerful_shadow} shown, \codename is also effective to reduce the attack accuracy of the attackers with more powerful shadow models. For instance, the attack accuracy of the Mlleaks attack drops to 50.01\% from 70.20\%, which means \codename is able to mitigate stronger attacks which having more information (e.g., an attacker has more data to train shadow models) than that we assumed in the main paper.

\subsection{Inference about the reference data}

Involving an in-distribution reference dataset in the defense mechanism is common in the literature. 
For instance, MemGuard uses a reference set to train the defense classifier. Min-Max uses it to train the inference model. 
Similarly, our approach uses it to train the \codename. 
Unfortunately, little has been discussed on whether such reference dataset brings vulnerability for data inference attacks.
Assuming the reference data are considered as members, we present the inversion error and the inference accuracy (we consider NSH attack) on the reference set $D_2$ and the test set $D_3$ for each defense in Table~\ref{tb:attack_d2}. 
Results show that 
the inference accuracy does not increase on the reference set compared with the original training data of the target classifier. 
\codename can still preserve the defense effect against the adversarial model inversion attack and the membership inference attack.
However, there might be opposing views on whether such reference datasets should be considered as members.

\begin{table}[t]
	\centering
        \small
	\caption{Results of model inversion attack and membership inference attack on the reference set for different defenses. The experiments are performed on the FaceScrub530 dataset.}
	\label{tb:attack_d2}
	\resizebox{\columnwidth}{!}{
	\begin{tabularx}{\columnwidth}{l|Y|Y}
		\hline
		Defense & Inversion error & Inference accuracy\\
		\hline
		Purifier & 0.0435 & 51.23\%\\
		\hline
		Min-Max & 0.0222 & 52.07\% \\
		\hline
		MemGuard & 0.0116 & 52.20\% \\
		\hline
	\end{tabularx}
	}
\end{table}

\subsection{Effect of \codename's Training Data}

\begin{table}[t]
	\centering
	\small
	\setlength{\tabcolsep}{6pt}
	\caption{Effect of the \codename's in-distribution training data on the 
		defense performance. The numbers are reported on the Purchase100 dataset.}
	\label{tb:effectreference}
	\resizebox{\columnwidth}{!}{
		\begin{tabularx}{\columnwidth}{c|c|Y|Y|Y}
			\hline
			Training set & Classification acc & NSH & 
			Mlleaks & Adaptive  \\
			\hline
			$D_2$ (5,000) & 83.85\% &  52.63\% & 50.09\% & 
			50.62\%   \\
			\hline
			$D_2$ (10,000) & 83.47\%  &  51.72\% & 50.12\% & 
			50.14\% \\
			\hline
			$D_2$ (20,000) & 83.23\% &  51.71\% & 50.09\% & 
			50.13\%  \\
			\hline
			$D_2$ (40,000) & 81.39\%  &  51.93\% & 50.09\% & 
			50.20\%  \\
			\hline
			$D_2$ (60,000) & 81.75\% &  51.84\% & 50.09\% & 
			50.10\%  \\
			\hline
			$D_1$ (20,000) & 84.23\%  &  52.29\% & 50.12\% & 
			50.07\%  \\
			\hline
		\end{tabularx}
	}
\end{table}

We investigate the effect of the \codename's training data by using different 
in/out-distribution data to train \codename.
Specifically, for in-distribution data, we vary the size of $D_2$ and also 
replace $D_2$ with $D_1$. For out-of-distribution data, we use CIFAR10 data to 
train the \codename for the FaceScrub530 classifier, and use randomly generated 
data to train the \codename for the Purchase100 classifier.

We present the effect of the in-distribution training data in 
Table~\ref{tb:effectreference}.
{
The results show that \codename is still effective. The membership inference accuracy (Mlleaks and Adaptive)  
is reduced to nearly 50\% regardless of the size of $D_2$. \codename is also insensitive to the size of the $D_2$. The difference of the defense performance is negligible as the size of $D_2$ changes from 5,000 to 60,000. This is good for the defender, as one can achieve good performance with a small reference set. However, when the size of $D_2$ becomes too large (i.e., 40,000 to 60,000), the classification accuracy drops to a certain extent. The reason could be that \codename starts to learn the detailed information of the confidence score vectors. As a result, the purified confidence score vector no longer concentrates on general 
patterns but becomes an accurate reconstruction, which hinders the classification utility.

When we use the classifier's training data $D_1$ to train the \codename, the defense performance is comparable to the ones on $D_2$. For example, the attack accuracy of the NSH attack is 52.29\%, which is marginally higher than the result of 51.71\% on $D_2$, but still acceptable.

}

Table~\ref{tb:effectoutof} shows the effect of the out-of-distribution training 
data. \codename can still mitigate the attacks, but at the cost of 
sacrificing the utility of the target classifier significantly. 
This is not surprising because \codename cannot extract useful patterns from 
the confidence scores on out-of-distribution data, which makes the purified 
confidence information meaningless.

\begin{figure}[t]
	\centering   
	\subfigure{
		\includegraphics[width=1\linewidth]{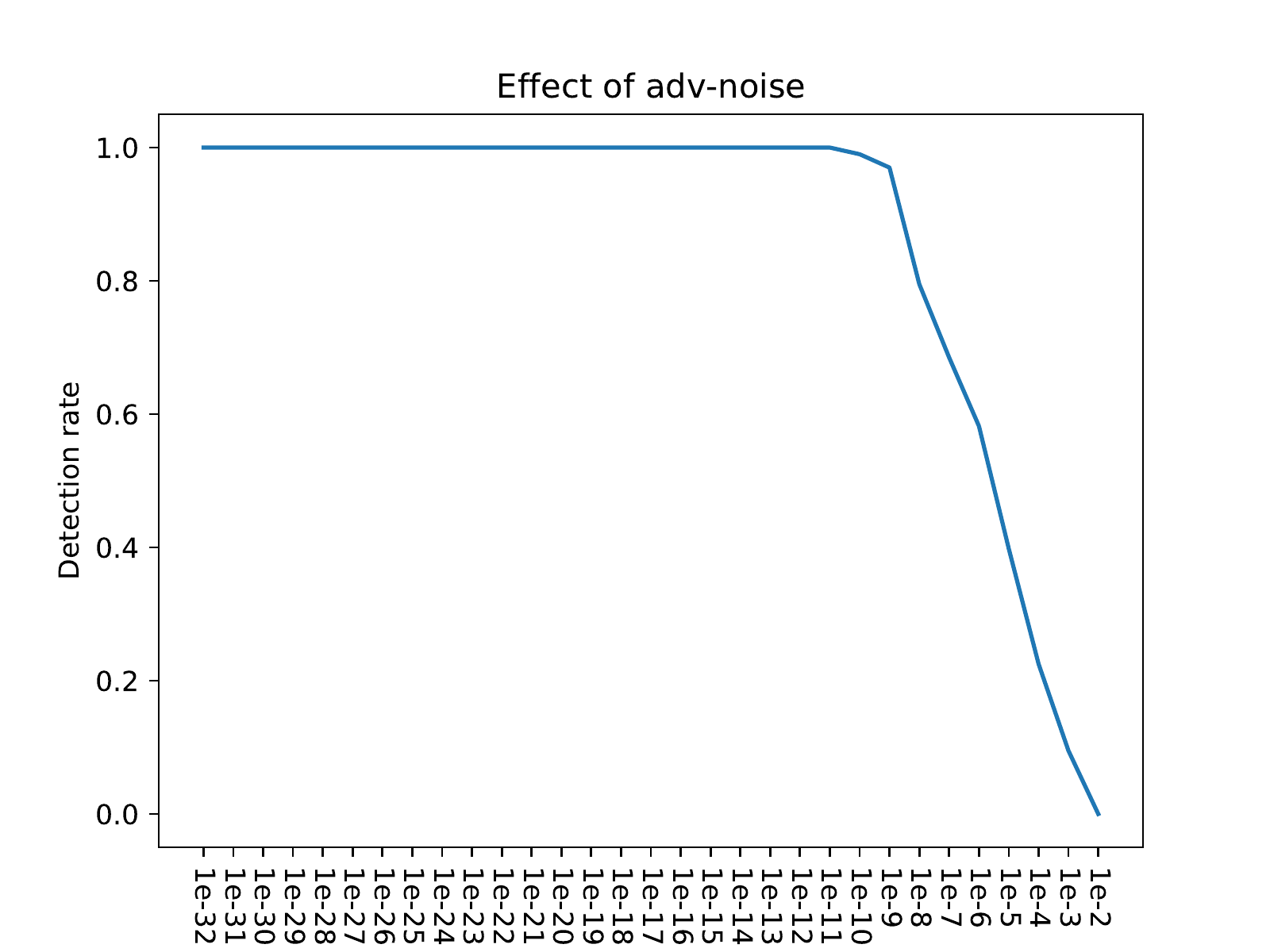}
	}
	\caption{
 The proportion of noise data that \codename can detect under the 
 FGSM attacks on the FaceScrub530 dataset.
  }
	\label{fig:near_member}
\end{figure}

\begin{table}[h]
	\centering
        \small
	\setlength{\tabcolsep}{1pt}
	\caption{Effect of the \codename's out-of-distribution training data on the 
		defense performance.}
	\label{tb:effectoutof}
	\resizebox{\columnwidth}{!}{
		\begin{tabularx}{\columnwidth}{l|l|Y|c|c|c}
			\hline
			Classifier & Purifier & Classification acc & NSH & 
			Mlleaks & Adaptive \\
			\hline
			FaceScrub530 & CIFAR10 & 40.05\% & 54.54\% & 50.11\% 
			& 50.50\% \\
			\hline
			Purchase100 & Random & 7.76\%  &  51.55\% & 50.20\% & 
			50.71\% \\
			\hline
		\end{tabularx}
	}
\end{table}

\subsection{Effectiveness of \codename
to detect noisy members}
We investigate the effectiveness of \codename
to detect noisy members, which means the members added noise by attackers intentionally. We use the adversarial attack methods (FGSM)\footnote{ICLR 2015 Explaining and Harnessing Adversarial Examples}
to create noisy members.  
As shown in Figure~\ref{fig:near_member}  that \codename can accurately detect the members
with noise $\left \| \eta \right \|_{\infty}<1e-10$ on FaceScrub530 dataset.

\begin{table}[]
\caption{Ablation study on the Label Swapper.}
\label{tb:Ablation}
\begin{tabular}{c|c|c}
\toprule
\multirow{2}{*}{Dataset}                       & \multirow{2}{*}{Defense} & \multirow{2}{*}{Transfer Attack} \\
                                               &                          &                                             \\ \hline
\multicolumn{1}{c|}{\multirow{3}{*}{CIFAR10}} & None                     & 0.5048                                      \\ \cline{2-3} 
\multicolumn{1}{c|}{}                         & CVAE                     & 0.5045                                      \\ \cline{2-3} 
\multicolumn{1}{c|}{}                         & CVAE + Label Swapper     & 0.4974                                     \\ \hline
\multirow{3}{*}{CIFAR100}                      & None                     & 0.6668                                      \\ \cline{2-3} 
                                               & CVAE                     &  0.6632                                      \\ \cline{2-3} 
                                               & CVAE + Label Swapper     & 0.4975                                      \\ \bottomrule
\end{tabular}
\end{table}

\subsection{Ablation study on the Label Swapper}
We conduct the ablation study on the label swapper to investigate the effectiveness of it in our defense mechanism. Given our design of label swapper to mitigate the attackers using only the hard labels to perform attacks, we verify its efficiency under label only transfer attack. Table~\ref{tb:Ablation} shows a significant AUC drop in label only transfer attack with the help of label swapper compared by those mechanisms without it. Additionally, the result indicates that label only transfer attack can obtain approximately the same AUC among models without any defense and \codename without the label swapper, which means \codename without label swapper fails to mitigate the label only attack as we have assumed. In summary, the label swapper plays an indelible role in PURIFIER to mitigate the label only attacks.

\subsection{Cost of \codename in Real-world Setting}
We discuss the latency and the memory cost introduced by \codename at this section. We report that the training time of the CIFAR10 model is 3.18h, and the extra cost introduced by \codename in it is 1.34h. Compared with 71.30h SELENA has introduced, \codename contributes to less latency time. Besides the training cost, \codename will spend extra 282.18s to predict 50k images in the inference time. We speculate that the latency of \codename is mostly caused by $k$NN. Hence we expect to replace $k$NN by a more efficient component in the future work. For the memory cost, \codename costs extra space for swap set. The sizes of it with different datasets are respectively 78.23K for CIFAR10, 1.14M for Purchase and 13.5M for Facescrub530. We believe the extra memory cost by \codename is negligible compared to the original training set.

\end{document}